\documentclass[manuscript,screen]{acmart}
\usepackage{algorithm}
\usepackage{algorithmic}
\usepackage{bibentry}
\usepackage{amsfonts}
\usepackage{subfigure}
\usepackage{multirow}
\usepackage{amsmath}

\usepackage{graphicx}
\usepackage{epsfig}
\usepackage{color}
\usepackage{epstopdf}
\usepackage{rotating}
\usepackage{caption}
\usepackage{float}
\usepackage{multicol}
\usepackage{amssymb}
\usepackage{graphicx}
\usepackage{bbding}
\usepackage{algorithm}
\usepackage{algorithmic}
\usepackage{balance}
\usepackage{microtype}
\usepackage{graphicx}
\usepackage{subfigure}
\usepackage{booktabs}
\usepackage{multirow}
\usepackage{amsmath}
\usepackage{mathtools}
\usepackage{etoolbox}
\usepackage{cases}
\usepackage{enumitem}
\usepackage{xcolor}
\usepackage{subfigure}
\usepackage{cases}

\newcommand{\cyan}[1]{{\color{cyan}#1}} % comments
\newcommand{\purple}[1]{{\color{purple}{#1}}} % comments

\usepackage{tikz}

\newcommand*\halfcirc[1][1ex]{%
	\begin{tikzpicture}
		\draw[fill] (0,0)-- (90:#1) arc (90:270:#1) -- cycle;
		\draw (0,0) circle (#1);
\end{tikzpicture}}
\newcommand*\fullcirc[1][1ex]{\tikz\fill (0,0) circle (#1);}

\AtBeginDocument{%
  }

\setcopyright{acmlicensed}
\copyrightyear{2024}
\acmYear{2024}
\acmDOI{XXXXXXX.XXXXXXX}

\acmConference[Conference acronym 'XX]{Make sure to enter the correct
  conference title from your rights confirmation emai}{June 03--05,
  2024}{Woodstock, NY}
\acmISBN{978-1-4503-XXXX-X/18/06}

\begin{document}

%%
%% The "title" command has an optional parameter,
%% allowing the author to define a "short title" to be used in page headers.
\title{A Survey of Mamba}

\author{Haohao Qu}
\email{haohao.qu@connect.polyu.hk}
\orcid{0000-0001-7129-8586}
\affiliation{%
  \institution{The Hong Kong Polytechnic University}
  \city{Hong Kong}
  \country{Hong Kong}
}

\author{Liangbo Ning}
\email{liangbo1123.ning@connect.polyu.hk}
\orcid{0000-0001-6903-8996}
\affiliation{%
  \institution{The Hong Kong Polytechnic University}
  \city{Hong Kong}
  \country{Hong Kong}
}

\author{Rui An}
\email{rui77.ang@connect.polyu.hk}
\orcid{}
\affiliation{%
  \institution{The Hong Kong Polytechnic University}
  \city{Hong Kong}
  \country{Hong Kong}
}

\author{Wenqi Fan}
\authornote{Corresponding author: Wenqi Fan, Department of Computing (COMP) \& Department of Management and Marketing (MM). }
\email{wenqifan03@gmail.com}
\orcid{ }
\affiliation{%
  \institution{The Hong Kong Polytechnic University}
  \city{Hong Kong}
  \country{Hong Kong}
}

\author{Tyler Derr}
\email{Tyler.Derr@vanderbilt.edu}
\orcid{0000-0002-0080-5998}
\affiliation{%
	\institution{Vanderbilt University}
	\city{Nashville}
	\country{USA}
}

\author{Hui Liu}
\orcid{0000-0002-3555-3495}
\affiliation{%
  \institution{Michigan State University}
  \country{USA}}
\email{liuhui7@msu.edu}

\author{Xin Xu}
\email{xin.xu@polyu.edu.hk}
\orcid{0000-0001-6143-6471}
\affiliation{%
  \institution{The Hong Kong Polytechnic University}
  \city{Hong Kong}
  \country{Hong Kong}
}

\author{Qing Li}
\email{qing-prof.li@polyu.edu.hk}
\orcid{0000-0003-3370-471X}
\affiliation{%
  \institution{The Hong Kong Polytechnic University}
  \city{Hong Kong}
  \country{Hong Kong}
}

\renewcommand{\shortauthors}{Qu et al.}

%%
%% The abstract is a short summary of the work to be presented in the
%% article.
\begin{abstract}
Deep learning (DL), as a vital technique, has sparked a notable revolution in artificial intelligence (AI), resulting in a great change in human lifestyles.
As one of the most representative DL techniques, the Transformer architecture has empowered numerous advanced models, especially the large language models (LLMs) that comprise billions of parameters, becoming a cornerstone in deep learning.
Despite the impressive achievements, Transformers still face inherent limitations, particularly the time-consuming inference resulting from the quadratic computation complexity of attention calculation. 
Recently, a novel architecture named \textbf{Mamba}, drawing inspiration from classical state space models (SSMs), has emerged as a promising alternative for building foundation models, delivering comparable modeling abilities to Transformers while preserving near-linear scalability concerning sequence length. 
This has sparked an increasing number of studies actively exploring Mamba's potential to achieve impressive performance across diverse domains.
Given such rapid evolution, there is a critical need for a systematic review that consolidates existing Mamba-empowered models, offering a comprehensive understanding of this emerging model architecture.
In this survey, we therefore conduct an in-depth investigation of recent Mamba-associated studies, covering three main aspects: \emph{the advancements of Mamba-based models}, \emph{the techniques of adapting Mamba to diverse data}, and \emph{the applications where Mamba can excel}.
Specifically, we first review the foundational knowledge of various representative deep learning models and the details of Mamba-1\&2 as preliminaries. 
Then, to showcase the significance of Mamba for AI, we comprehensively review the related studies focusing on Mamba models' architecture design, data adaptability, and applications.
Finally, we present a discussion of current limitations and explore various promising research directions to provide deeper insights for future investigations. 

\end{abstract}

%%
%% The code below is generated by the tool at http://dl.acm.org/ccs.cfm.
%% Please copy and paste the code instead of the example below.
%%

% \ccsdesc[300]{Computing methodologies~Neural Networks}
% \ccsdesc{Computing methodologies~Neural Networks}
% \ccsdesc[100]{Computing methodologies~Neural Networks}

%%
%% Keywords. The author(s) should pick words that accurately describe
%% the work being presented. Separate the keywords with commas.

% \received{20 February 2007}
% \received[revised]{12 March 2009}
% \received[accepted]{5 June 2009}

%%
%% This command processes the author and affiliation and title
%% information and builds the first part of the formatted document.
\maketitle

\section{Introduction}
\label{se:intro}
Over the past two decades, deep learning (DL), as the most prominent artificial intelligence (AI) technique, has brought about a revolution in various domains such as healthcare~\cite{jones2024causal}, autonomous systems~\cite{guan2024world,fan20244d}, recommender systems~\cite{li2024embedding,zhao2024recommender}, and financial services~\cite{prata2024lob,zhang2024deep}. This period has witnessed the emergence of numerous deep neural networks (DNNs) that have significantly altered human lifestyles, offering immense convenience to individuals.
One notable example is U-Net~\cite{ronneberger2015u,si2024freeu}, a robust deep learning model within the field of vision, which is extensively employed in medical imaging for the examination of radiology scans like MRI and CT scans.
Its application assists in the identification and diagnosis of diseases, showcasing its effectiveness in this critical healthcare domain~\cite{williams2024unified,lin2022ds}. 
Moreover, Graph Neural Networks (GNNs) are employed in handling graph-structured data to support smart services, such as recommender systems that suggest personalized content, products, or services to users~\cite{fan2020graph,fan2019graph,wu2019session}.
Furthermore, Recurrent Neural Networks (RNNs) are extensively adopted in machine translation due to their ability to capture the sequential and contextual information essential for accurate translations~\cite{liu2014recursive,su2017lattice}, empowering individuals from diverse linguistic backgrounds to effectively communicate and comprehend each other's ideas, opinions, and information. 

Among the various DL architectures, Transformers have recently stood out and established their dominance across a broad spectrum of applications~\cite{dong2023towards,vert2023will}. 
For instance, as the most representative large foundation models, large language models (LLMs) like ChatGPT and GPT4 are fundamentally built on the Transformer architecture~\cite{achiam2023gpt,qu2024tokenrec,zhao2024recommender}. 
By scaling their model sizes to billions and training on a mix of diverse data sources, these Transformer-based models have demonstrated human-level intelligence with their impressive capabilities in language understanding, common sense reasoning, and in-context learning~\cite {zhang2023survey,fan2024graph}. 
This remarkable success is bolstered by the attention mechanism~\cite{vaswani2017attention}, which enables the Transformer-based models to concentrate on relevant parts of the input sequence and facilitate better contextual understanding.
However, the attention mechanism also introduces a significant computational overhead that increases quadratically with the input size~\cite{lu2021soft,zhu2021long}, which presents challenges in processing lengthy inputs.
For example, the rapid growth in computational cost makes Transformers impractical or infeasible to process substantial sequences, thereby limiting their applicability in tasks like document-level machine translation~\cite{maruf2021survey} or long document summarization~\cite{koh2022empirical}.

\begin{figure}[h]
\centering
\includegraphics[width=\linewidth]{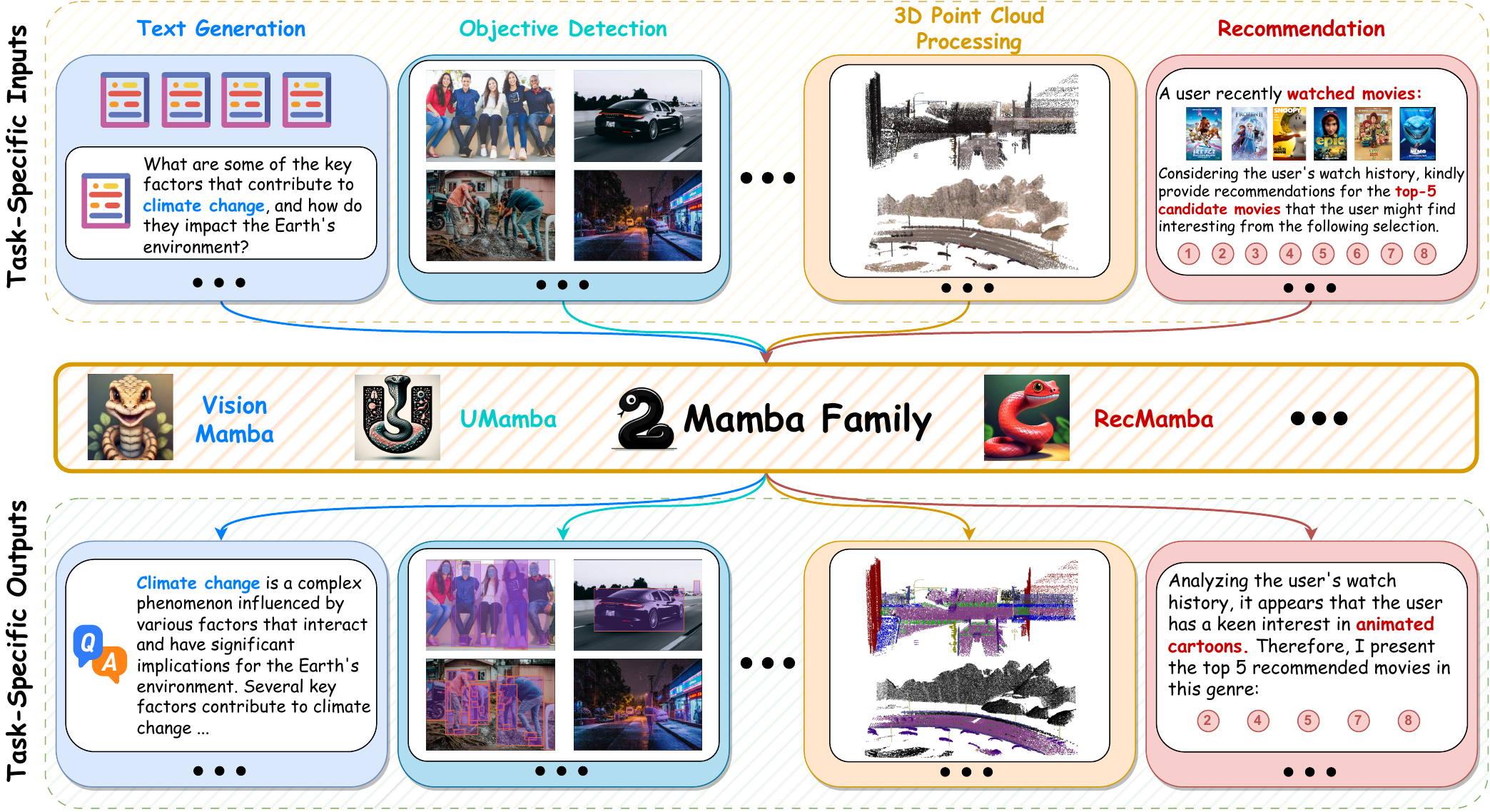}
\caption{Examples of the applications of Mamba-based models for different downstream tasks.}
\label{fig:intro}
\end{figure}

Recently, a promising architecture, structured state space sequence models (SSMs)~\cite{gu2021efficiently}, has emerged to efficiently capture complex dependencies in sequential data, becoming a formidable competitor to the Transformer. 
These models, inspired by classical state space models~\cite{kalman1960new}, can be considered a fusion of recurrent neural networks and convolutional neural networks.
They can be computed efficiently using either recurrence or convolution operations, achieving linear or near-linear scaling with sequence length, thus significantly reducing the computational costs. 
More specifically, as one of the most successful SSM variants, \textbf{Mamba} achieves comparable modeling capabilities to Transformers while maintaining linear scalability with sequence length~\cite{gu2023mamba}, propelling it into the realm of focal topics.
Mamba first introduces a simple yet effective selection mechanism that enables the model to filter out irrelevant information while retaining necessary and relevant data indefinitely by parameterizing the SSM parameters based on the input. 
Then, Mamba proposes a hardware-aware algorithm to compute the model recurrently with a scan instead of a convolution, achieving up to 3$\times$faster computation on A100 GPUs. 
As shown in Figure~\ref{fig:intro}, the powerful modeling capabilities for complex and lengthy sequential data, along with near-linear scalability, position Mamba as an emerging foundation model, poised to revolutionize various domains of research and applications, such as computer vision~\cite{xu2024survey,zhu2024vision}, natural language processing~\cite{lieber2024jamba,zhao2024cobra}, healthcare~\cite{ruan2024vm,xing2024segmamba,wang2024mamba}, etc. 
For example, \citet{zhu2024vision} propose Vim, which is 2.8$\times$faster than DeiT~\cite{touvron2021training} and saves 86.8\% GPU memory when extracting features for high-resolution images. 
\citet{mamba2} show the connections between SSMs and variants of attention and propose a new architecture that refines selective SSM, achieving 2-8$\times$ faster on language modeling. 

Motivated by the powerful long-sequence modeling capabilities of Mamba and its great efficiency, a substantial body of literature has emerged, focusing on employing and improving Mamba on various downstream tasks.
Given this significant surge in studies related to Mamba, it is crucial to conduct a comprehensive review of existing literature and deliberate on potential directions for future research. 
In this survey, we thus conduct a comprehensive review of Mamba from several perspectives to provide newcomers with a fundamental understanding of Mamba's inner workings while helping experienced practitioners stay abreast of its latest developments.
Specifically, the remaining survey is organized as follows: In Section~\ref{se:pre}, we recall the background knowledge of various representative deep neural networks, including RNNs, Transformers, and State Space Models, while the details of Mamba are introduced in Section~\ref{se:mamba}.
Subsequently, we summarize the recent advancements in Mamba-based studies from the perspectives of block design, scanning mode, and memory management in Section~\ref{se:module}. 
Then, Section~\ref{se:data} presents the techniques of adapting Mamba to diverse data, including sequential and non-sequential data.
Besides, representative applications of Mamba models are introduced in Section~\ref{se:app}, while the challenges and future directions are presented in Section~\ref{se:future_direction}.
Finally, we summarize the whole survey in Section~\ref{se:conclusion}.

Concurrent with our survey, several related surveys have been released, purely focusing on state space models~\cite{patro2024mamba,wang2024state} and Vision Mamba~\cite{zhang2024survey,liu2024vision,xu2024survey}. 
Diverging from these surveys, this paper is centered on the associated research concerning Mamba.
It systematically analyzes existing literature from a novel standpoint to explore \emph{the evolution of Mamba architecture} and \emph{the data adaptation methods utilized in Mamba-based models}.

\section{Preliminary}
\label{se:pre}
Mamba is deeply intertwined with the recurrent framework of Recurrent Neural Networks (RNNs), the parallel computation and attention mechanism of Transformers, and the linear property of State Space Models (SSMs).
Therefore, this section aims to present an overview of these three prominent architectures.

\subsection{Recurrent Neural Networks (RNNs)}
RNNs excel in processing sequential data due to their capability to retain internal memory~\cite{graves2012long}.
Such networks have demonstrated remarkable effectiveness in a multitude of tasks that involve analyzing and predicting sequences, e.g., speech recognition, machine translation, natural language processing, and time-series analysis~\cite{sutskever2011generating,hermans2013training}.
In order to grasp the foundations of recurrent models, this section will offer a brief overview of the standard RNN formulation.

Specifically, at each discrete time step $k$, the standard RNN specifically processes a vector $x_k \in \mathbb{R}^D$ along with the previous step's hidden state $h_{k-1} \in \mathbb{R}^N$ to produce an output vector $o_k \in \mathbb{R}^O$ and update the hidden state to $h_k \in \mathbb{R}^N$.
The hidden state serves as the network's memory and retains information about the past inputs it has seen.
This dynamic memory allows RNNs to process sequences of varying lengths.
Formally, it can be written as
\begin{align}
\label{eq:rnn}
    h_k &= \text{tanh}(\mathbf{W}_{hx} x_k + \mathbf{W}_{hh} h_{k-1} + b_h), \\
    o_k &= \mathbf{W}_{oh} h_k + b_o,
\label{eq:rnn2}
\end{align}
where $\mathbf{W}_{hx} \in \mathbb{R}^{N \times D}$ is the weight matrix responsible for processing model inputs into hidden states, $\mathbf{W}_{hh} \in \mathbb{R}^{N \times N}$ is the recurrent connections between hidden states, $\mathbf{W}_{oh} \in \mathbb{R}^{O \times N}$ represents the weight used to generate outputs derived from hidden states, $b_h \in \mathbb{R}^N$ and $b_o \in \mathbb{R}^O$ correspond the biases, and $\text{tanh}$ denotes the hyperbolic tangent activation function introducing non-linearity to the RNN model.
In other words, RNNs are nonlinear recurrent models that effectively capture temporal patterns by harnessing the historical knowledge stored in hidden states.

However, there are several limitations associated with RNNs.
First, RNNs have a restricted capability to effectively extract long-range dynamics within input sequences.
As information traverses through successive time steps, the repeated multiplication of weights in the network can lead to dilution or loss of information.
Consequently, it becomes challenging for RNNs to retain and recall information from earlier time steps while making predictions.
Second, RNNs process sequential data incrementally, restricting their computational efficiency since each time step relies on the preceding one.
This makes parallel computations challenging for them.
Furthermore, conventional RNNs lack built-in attention mechanisms, which allow the network to capture global information within input sequences.
This absence of attention mechanisms hinders the network's ability to selectively model the crucial segments of the data.
To overcome these constraints, Transformers and State Space Models have emerged, each tackling these challenges from different perspectives.
These two approaches will be further elaborated upon in the subsequent subsections.

\subsection{Transformers}
The Transformer~\cite{vaswani2017attention} is a groundbreaking model in the realm of deep learning, revolutionizing various AI applications. 
Its introduction marked a significant departure from traditional sequence-to-sequence models by employing a \textbf{self-attention} mechanism, facilitating the capture of global dependencies within model inputs. 
For instance, in natural language processing, this self-attention capability allows the model to comprehend relationships between various positions in a sequence. 
It achieves this by assigning weights to each position based on its significance relative to other positions. 
More specifically, a sequence of input vectors $\textbf{x}$ is first transformed into three types of vectors: \textbf{Query $Q$}, \textbf{Key $K$}, and \textbf{Value $V$} by utilizing linear transformations of the original input, defined by: 
\begin{align}
    Q=\textbf{x}\cdot \mathbf{W}^Q, K=\textbf{x}\cdot \mathbf{W}^K, V=\textbf{x}\cdot \mathbf{W}^V, 
\end{align}
where $\mathbf{W}^Q$, $\mathbf{W}^K$, and $\mathbf{W}^V$ are the trainable parameters. 
The attention scores are computed by calculating the dot product of $Q$ and $K$, then scaling the result by $\sqrt{d_K}$, where $d_K$ is the dimension of the key vectors. 
Such procedures are then passed through a Softmax function to normalize the scores $S$ and produce attention weights, defined by: 
\begin{align}
S = \text{Softmax}(\frac{QK^T}{\sqrt{d_K}})V ,
\label{eq:self_attention}
\end{align}

Apart from performing a single attention function, multi-head attention is introduced to enhance the model's ability to capture different types of relationships and provide multiple perspectives on the input sequence. 
In multi-head attention, an input sequence is processed in parallel by multiple sets of self-attention modules. 
Each head operates independently, performing the exact computations as in the standard self-attention mechanism. 
The attention weights from each head are then combined to create a weighted sum of the value vectors. 
This aggregation step allows the model to leverage information from multiple heads and capture diverse patterns and relationships in the input sequence. 
Mathematically, the multi-head attention is computed as follows: 
\begin{align}
\begin{split}
    \text{MultiHead}&(Q,K,V) =(S_1\oplus S_2 \oplus ... \oplus S_m) \cdot \mathbf{W}^O, \\
      \text{where}~~~~ & S_i = \text{Softmax}(\frac{Q_i K_i^T}{\sqrt{d_K}})V_i, ~~i \in [1,m], 
\end{split}
\end{align}
where $m$ is the number of attention heads, $ \oplus$ is the concatenation operation, and $\mathbf{W}^O$ is the linear transformation to project the multi-head attention scores to the final values. 

\subsection{State Space Models}
State Space Models (SSMs) are a traditional mathematical framework utilized to depict the dynamic behavior of a system over time~\cite{kalman1960new}.
Recent years have found the widespread applications of SSMs in diverse fields like control theory, robotics, and economics~\cite{gu2021combining,gu2021efficiently}.
At its core, SSMs embody the system's behavior through a collection of hidden variables referred to as "states", enabling it to capture temporal data dependencies effectively.
Different from RNNs, SSMs are linear models characterized by their associative properties.
To be specific, in a classical state space model, two fundamental equations are formulated, i.e., state equation and observation equation, 
to model the relationships between input $x(t) \in \mathbb{R}$ and output $y(t) \in \mathbb{R}$ at current time $t$, through a N-dimensional hidden state $h(t) \in \mathbb{R}^N$.
The process can be written by
\begin{align}
\label{eq:ssm1}
    h'(t) & = \mathbf{A}h(t) + \mathbf{B}x(t), \\
    y(t) & = \mathbf{C}h(t) + \mathbf{Z}x(t),
\label{eq:ssm2}
\end{align}
where $h'(t)$ is the derivative of current state $h(t)$, $\mathbf{A} \in \mathbb{R}^{N \times N}$ is the state transition matrix that describes how states change over time, $\mathbf{B} \in \mathbb{R}^{N \times 1}$ is the input matrix that controls how inputs affect state changes, $\mathbf{C} \in \mathbb{R}^{1 \times N}$ denotes the output matrix that indicates how outputs are generated based on current states, and {$\mathbf{Z} \in \mathbb{R}$} represents the command coefficient that determines how inputs affect outputs directly.
In general, most SSMs exclude the second term in the observation equation, i.e., set $\mathbf{Z}x(t) = 0$, which can be recognized as a skip connection in deep learning models.
{Broadly speaking, given an input $x(t)$ with $D$ dimensions, the SSM computation will be calculated separately for each dimension to produce a $D$-dimensional output $y(t)$.
In this case, the input matrix $\mathbf{B} \in \mathbb{R}^{N \times D}$, the output matrix $\mathbf{C} \in \mathbb{R}^{D \times N}$, and the command matrix $\mathbf{Z} \in \mathbb{R}^{D}$ (if available), while the state transition matrix remains unchanged, i.e., $\mathbf{A} \in \mathbb{R}^{N \times N}$.}

\subsubsection{\textbf{Discretization}}
To adhere to the requirements of machine learning settings for various real-world scenarios, SSMs must undergo a process of discretization that transforms continuous parameters into discrete parameters.
Discretization methods generally aim to partition continuous time into $K$ discrete intervals with as equal an integration area as possible.
To achieve the goal, as one of the most representative solutions, 
Zero-Order Hold (ZOH)~\cite{zhang2007comparison,pechlivanidou2022zero} is successfully employed in SSMs, which assumes that the function value remains constant over the interval $\Delta = [t_{k-1}, t_k]$.
After ZOH discretization, the SSM equations can be rewritten as
\begin{align}
    h_k &= \overline{\mathbf{A}}h_{k-1} + \overline{\mathbf{B}}x_k, \\
    y_k &= \mathbf{C}h_k, 
\end{align}
where $\overline{\mathbf{A}}=\exp (\Delta \mathbf{A})$, and $\overline{\mathbf{B}} = (\Delta \mathbf{A})^{-1} (\exp(\Delta \mathbf{A})-\mathbf{I}) \cdot \Delta \mathbf{B}$, $k$ is the discrete time step.
From these formulas, it is clear that the discrete SSM has a similar structure to recurrent neural networks and, therefore, discrete SSMs can accomplish inference processes with higher efficiency compared to Transformer-based models that compute attention on all inputs in each auto-regressive decoding iteration.

\subsubsection{\textbf{Convolutional Computation}}
The discrete SSM, being a linear system, possesses the associated property and, therefore, integrates seamlessly with convolutional computation.
More specifically, it can calculate the output at each time step independently as follows:
\begin{align}
    y_0 &= \mathbf{C} \overline{\textbf{A}}^0 \overline{\mathbf{B}} x_0, \\
    y_1 &= \mathbf{C} \overline{\textbf{A}}^1 \overline{\mathbf{B}} x_0 + \mathbf{C} \overline{\textbf{A}}^0 \overline{\mathbf{B}} x_1, \\
    y_2 &= \mathbf{C} \overline{\textbf{A}}^2 \overline{\mathbf{B}} x_0 + \mathbf{C} \overline{\textbf{A}}^1 \overline{\mathbf{B}} x_1 + \mathbf{C} \overline{\textbf{A}}^0 \overline{\mathbf{B}} x_2, \\
    & ...... \\
    y_k &= \mathbf{C} \overline{\textbf{A}}^k \overline{\mathbf{B}} x_0 + \mathbf{C} \overline{\textbf{A}}^{k-1} \overline{\mathbf{B}} x_1 + ... + \mathbf{C} \overline{\textbf{A}}^1 \overline{\mathbf{B}} x_{k-1} + \mathbf{C} \overline{\textbf{A}}^0 \overline{\mathbf{B}} x_k.
\end{align}

By creating a set of convolutional kernels $\overline{\mathbf{K}} = (\mathbf{C} \overline{\mathbf{B}},...,\mathbf{C} \overline{\mathbf{A}}^k \overline{\mathbf{B}}, ...)$, the recurrent computation can be converted to a convolutional form as: 
\begin{align}
    \mathbf{y} = \mathbf{x} * \overline{\mathbf{K}},
\end{align}
where $\mathbf{x} = [x_0, x_1, ...]$ and $\mathbf{y} = [y_0, y_1, ...] \in \mathbb{R}^L$ denote the input and output sequences, respectively, while $L$ is the sequence length.
This convolutional computation allows SSMs to take full advantage of modern matrix computation hardware (e.g., GPUs) to enable parallel computing during the training process, which is impossible with RNNs utilizing nonlinear activation functions.

\subsubsection{\textbf{Relationship among RNN, Transformer, and SSM}}
\begin{figure*}[t]
\centering
{\includegraphics[width=0.95\linewidth]{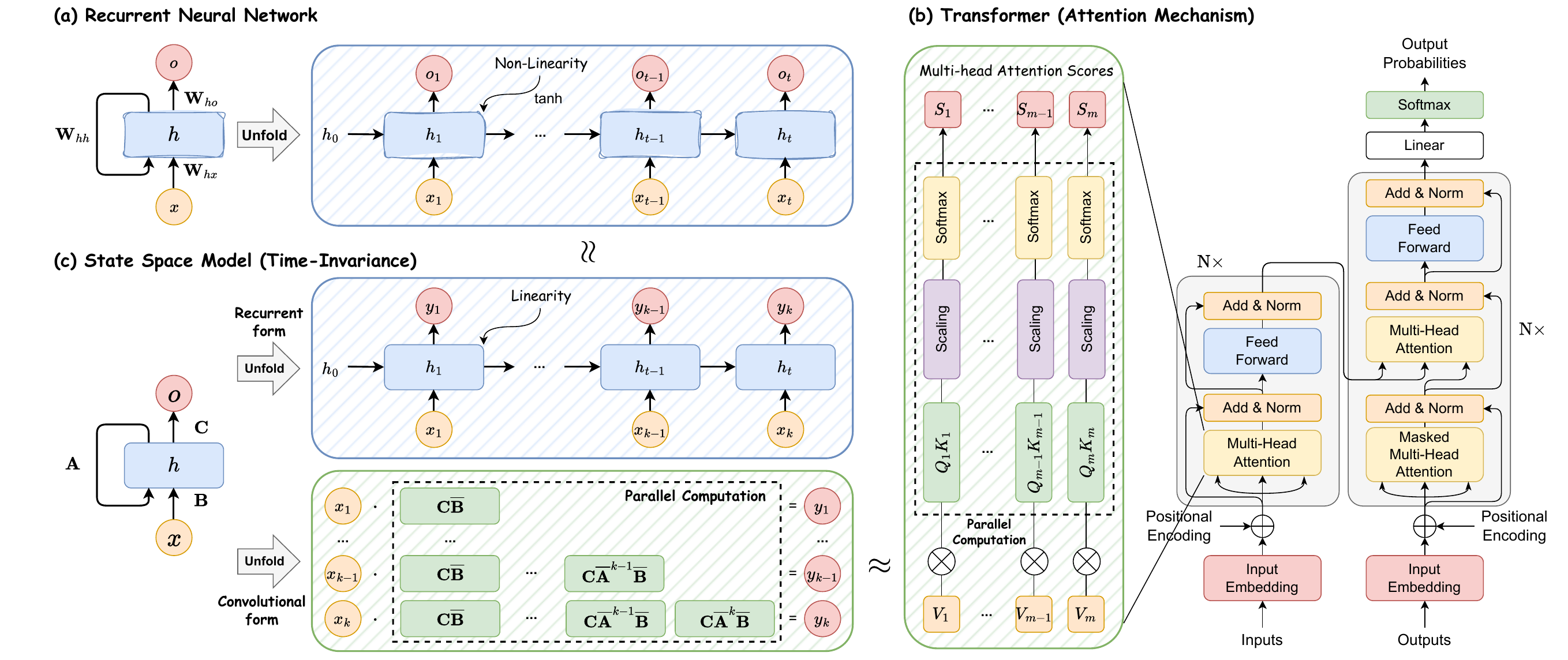}}
\caption{An illustration of representative model architectures, namely Recurrent Neural Network (RNN), Transformer, and State Space Model (SSM). (a) RNNs function within a nonlinear recurrent framework, facilitating rapid outputs during auto-regressive inference. (b) Transformers execute matrix multiplications concurrently across numerous query-key pairs, facilitating parallel training. (c) SSMs exhibit versatility by accommodating both recurrent and convolutional computations due to their linear nature.
This fusion harnesses the strengths of RNNs and Transformers, allowing SSMs for recurrent inference and parallel training.
Despite this, traditional time-invariant SSMs fall short in context-aware modeling, resulting in diminished performance in specific tasks.}
\label{fig:transformer_and_mamba}
\end{figure*}

The computation algorithms of RNN, Transformer, and SSM are depicted in Figure~\ref{fig:transformer_and_mamba}.
% RNN
On the one hand, the conventional RNN operates within a non-linear recurrent framework where each computation depends solely on the previous hidden state and the current input.
While this format allows RNNs to quickly generate outputs during auto-regressive inference, it hampers their ability to fully exploit GPU parallel computing, leading to slower model training.
% Transformer
On the other hand, the Transformer architecture performs matrix multiplications in parallel across multiple query-key pairs, which can be efficiently distributed across hardware resources, enabling faster training of attention-based models.
However, when it comes to generating responses or predictions from Transformer-based models, the inference process can be time-consuming.
For instance, the auto-regressive design of language models entails generating each token in the output sequence sequentially, which requires repetitive calculations of attention scores at each step, leading to slower inference times.
% SSM
As shown in Table~\ref{tab:comparison},  unlike RNNs and Transformers, which are limited to supporting only one type of computation, discrete SSMs have the flexibility to support both recurrent and convolutional computations, given their linear properties.
This unique capability allows SSMs to achieve not only efficient inference but also parallel training.
However, it should be noted that the most conventional SSMs are time-invariant, meaning that their $\mathbf{A}$, $\mathbf{B}$, $\mathbf{C}$, and $\Delta$ are unrelated to the model input $x$.
This would limit context-aware modeling, which leads to inferior performance of SSMs in certain tasks such as selective copying~\cite{gu2023mamba}. 

% add a table
\begin{table}[b]
  \centering
  \caption{{Pros and cons of three primary architectures-RNNs, Transformers, and SSMs-in auto-regressive sequential modeling tasks.}}
  \vskip -0.1in
    \begin{tabular}{clll}
    \toprule
    \textbf{Comparison} & \textbf{RNNs} & \textbf{Transformers} & \textbf{SSMs} \\
    \midrule
    Training Speed & \purple{Slow} (Recurrent) & \cyan{Fast} (Parallel) & \cyan{Fast} (Convolutional) \\
    Inference Speed & \cyan{Fast} (Recurrent)  & \purple{Slow} (Quadratic-Time)  & \cyan{Fast} (Recurrent) \\
    Complexity & $O(L D^2)$ & $O(L^2 D)$ & $O(L D^2)$ \\
    Modeling Capabilities & \halfcirc ~ (Hidden State) & \fullcirc ~ (Attention) & \halfcirc ~ (Time-Invariance) \\
    \bottomrule
    \end{tabular}%
  \label{tab:comparison}%
\end{table}%

\section{Mamba}
\label{se:mamba}
To address the aforementioned drawback of traditional SSMs in terms of their inferior context-aware capabilities, \textbf{Mamba} is proposed by ~\cite{gu2023mamba} as a potential alternative that promises to be a general sequence foundation model backbone.
More recently, \textbf{Mamba-2}~\cite{mamba2} proposes Structured Space-State Duality (SSD) that establishes a robust theoretical framework connecting structured SSMs and various forms of attention, allowing us to transfer algorithmic and systems optimizations originally developed for Transformers to SSMs.
In this section, we will give a concise and clear introduction to Mamba and Mamba-2. 

\subsection{Mamba-1: Selective State Space Model with Hardware-aware Algorithms}
\begin{figure}[b]
\centering
{\includegraphics[width=0.9\linewidth]{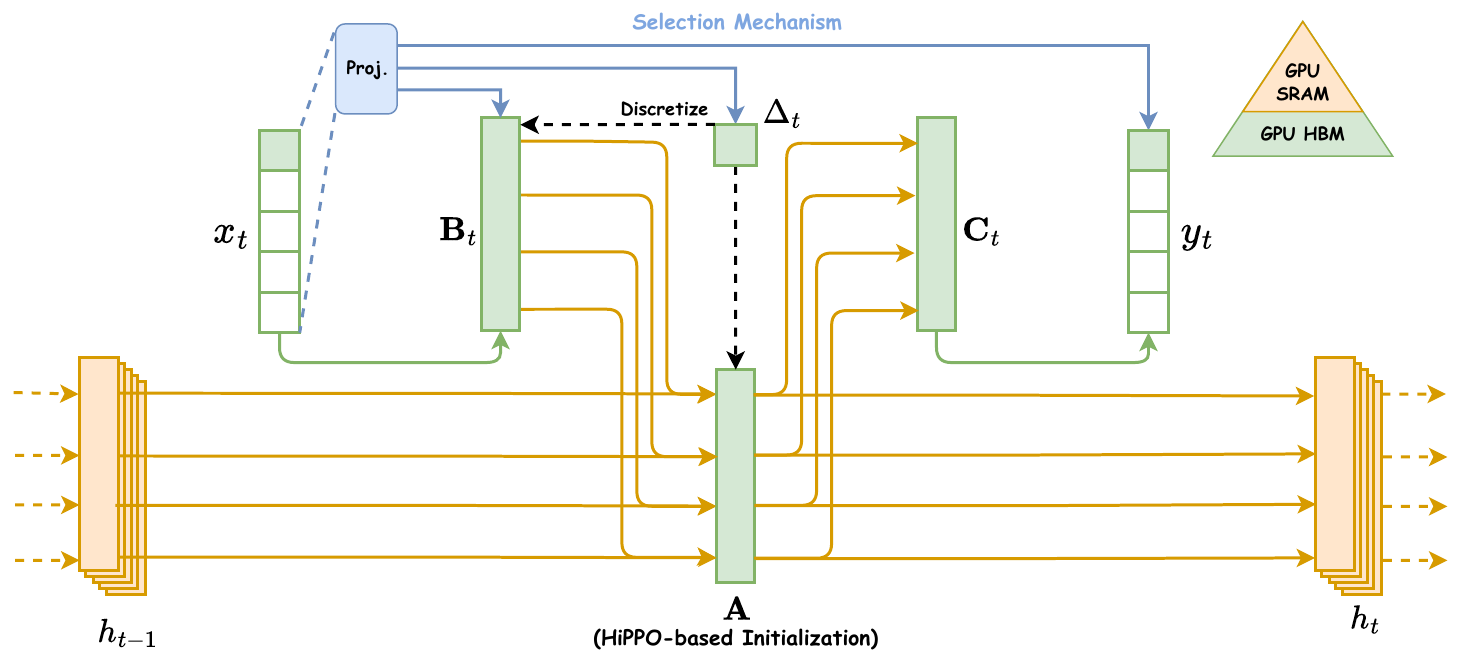}}
\vskip -0.1in
\caption{Overview of the Selective State Space Model with hardware-aware state expansions. The selective mechanism introduces input-dependent dynamics, while the hardware computation algorithm optimizes memory usage within the more efficient levels of GPU memory hierarchy.
}
\label{fig:mamba}
\end{figure}

Conventional SSMs have shown limited effectiveness in modeling text and other information-dense data, impeding their progress in deep learning.
In the pursuit of empowering SSMs with Transformers' modeling capabilities, \citet{gu2023mamba} introduce three innovative techniques based on Structured State Space Models, i.e., 
High-order Polynomial Projection
Operator (HiPPO)-based Memory Initialization, Selection Mechanism, and Hardware-aware Computation, as illustrated in Figure~\ref{fig:mamba}.
These techniques aim to enhance the capabilities of SSMs in long-range linear-time sequence modeling.
In particular, the initialization strategy establishes a coherent hidden state matrix, effectively facilitating long-range memory. 
Then, the Selection Mechanism empowers SSMs to acquire content-aware representations. 
Lastly, Mamba crafts two hardware-aware computation algorithms, Parallel Associative Scan and Memory Recomputation, to enhance training efficiency.

\subsubsection{\textbf{HiPPO-based Memory Initialization}}
Modeling and learning from sequential data represent foundational challenges in contemporary machine learning, forming the bedrock for various tasks, including language modeling, speech recognition, and video processing.
A fundamental component for modeling intricate and long-term temporal dependencies lies in memory, encompassing the ability to store and integrate information from preceding time steps~\cite{hu2017state}. 
Similar to RNNs, preserving and forgetting the historical hidden states (i.e., the matrix $\mathbf{A}$) play a critical role in SSMs to achieve satisfying performances.
In previous structured state space sequence models (SSMs), there have been suggestions for special initializations, especially in the case of complex-valued models.
These special initializations have proven beneficial in various scenarios, including situations with limited data availability.
Similarly, Mamba focuses primarily on the initialization of the hidden state matrix $\mathbf{A}$ to capture complex temporal dependencies.
This is accomplished through the utilization of the HiPPO theory~\cite{gu2020hippo} with an innovative scaled Legendre measure (LegS), ensuring careful consideration of the complete historical context rather than a limited sliding window.
To be specific, the HiPPO-LegS assigns uniform weight to all historical data points, which can be expressed as:
\begin{align}
    \mathbf{A}^\text{HiPPO}_{nk} = -
    \begin{cases}
    (2n+1)^{\frac{1}{2}} (2k+1)^{\frac{1}{2}} & \text{if} \ n > k \\
    n+1 & \text{if} \ n = k \\
    0 & \text{if} \ n < k
    \end{cases},
\end{align}
where $n$ is the number of polynomials, and $k$ denotes the particular discrete time steps.
Building upon the HiPPO theory, Mamba introduces two simple initialization methods for the complex and real cases, i.e., S4D-Lin and S4D-Real~\cite{gu2022parameterization}, as presented in
\begin{align}
    \mathbf{A}_{dn} = -
    \begin{cases}
    \frac{1}{2} - ni & \text{S4D-Lin} \\
    n+1 & \text{S4D-Real}
    \end{cases},
\end{align}
where $n$ is the $n$-th element of $\mathbf{A}$ for all input dimensions $d=1,2,...,D$.
Given such an initialization, the model can learn long-dependent memory that experiences smaller degradation of newer steps and larger degradation of older steps by compressing and reconstructing the input information signal.
According to the formulas, HiPPO-LegS possesses advantageous theoretical properties: it remains consistent across input timescales and offers rapid computation~\cite{gu2020hippo}.
Additionally, it has bounded gradients and approximation errors, facilitating the parameter learning process.

\subsubsection{\textbf{Selection Mechanism}}
Conventional state space models are unable to produce personalized outputs based on specific model inputs (i.e., the content-aware modeling ability) due to the property of Time Invariance.
To provide SSMs with a capability similar to that of attention mechanisms, Mamba designs a time-varying selection mechanism that parameterizes the weight matrices according to model inputs.
Such innovation empowers SSMs to filter out extraneous information while retaining pertinent details indefinitely.
Formally, the selection mechanism involves setting the interval $\Delta$, and matrices $\mathbf{B}$, $\mathbf{C}$ as functions of the input $\mathbf{x} \in \mathbb{R}^{B \times L \times D}$, which can be formulated as: 
\begin{align}
    \mathbf{B} \rightarrow \mathbf{S}^{\mathbf{B}} &= \mathbf{W}^{\mathbf{B}} \mathbf{x}, \\
    \mathbf{C} \rightarrow \mathbf{S}^{\mathbf{C}} &= \mathbf{W}^{\mathbf{C}} \mathbf{x}, \\
    \Delta \rightarrow \mathbf{S}^{\Delta} &= \tau_{\Delta} \cdot \text{BroadCast}_D(\mathbf{W}^{\Delta} \mathbf{x}),
\end{align}
where $\mathbf{S}^{\mathbf{B}} \in \mathbb{R}^{B \times L \times N}$, $\mathbf{S}^{\mathbf{C}} \in \mathbb{R}^{B \times L \times N}$, and $\mathbf{S}^{\Delta} \in \mathbb{R}^{B \times L \times D}$ are the selective space matrices that function of the input to achieve content-aware modeling. $B$, $L$, $D$, and $N$ represent the batch size, input length, input feature size, and hidden channel number, respectively.
{The activation function $(\tau_{\Delta} = \text{softplus})$ is utilized for $\Delta$.}
Notably, $\mathbf{W}^{\mathbf{B}} \in \mathbb{R}^{N \times D}$, $\mathbf{W}^{\mathbf{C}} \in \mathbb{R}^{N \times D}$, and $\mathbf{W}^{\Delta} \in \mathbb{R}^{D \times 1}$ are the selection weights (i.e., linear parameterized projections) for corresponding components, and $\text{BroadCast}_D$ means to broadcast the result to all the dimensions $d =1,2,..,D$.
Subsequently, the selective SSMs undergo discretization using a common statistical technique, Zero-Order Hold (ZOH)~\cite{pechlivanidou2022zero}, as presented in
\begin{align}
\overline{\mathbf{A}} \rightarrow \mathbf{S}^{\overline{\mathbf{A}}} & = \exp(\mathbf{S}^{\Delta} \mathbf{A}), \\
\overline{\mathbf{B}} \rightarrow    \mathbf{S}^{\overline{\mathbf{B}}} & = (\mathbf{S}^{\Delta} \mathbf{A})^{-1} (\exp(\mathbf{S}^{\Delta} \mathbf{A})-\mathbf{I}) \cdot \mathbf{S}^{\Delta} \mathbf{S}^{B},
\end{align}
where $\mathbf{S}^{\overline{\mathbf{A}}} \in \mathbb{R}^{B \times L \times D \times N}$ and $\mathbf{S}^{\overline{\mathbf{B}}} \in \mathbb{R}^{B \times L \times D \times N}$ are the selective state transition matrix and the input matrix, respectively, which become the functions of input $\mathbf{x}$.
By doing so, the discrete SSM has changed from time-invariant to time-varying (i.e., content-aware) as
\begin{align}
    \mathbf{y} = \text{SSM}(\mathbf{A}, \mathbf{B}, \mathbf{C})(\mathbf{x}),
\end{align}
which generates output $\mathbf{y} \in \mathbf{R}^{B \times L \times D}$ depending on the input $\mathbf{x}$.
Note that the time-varying selection mechanism in Mamba has a similar structure to the attention mechanism in Transformer, i.e., both perform operations based on inputs and their projections, which allows Mamba's SSM to achieve a flexible content-aware modeling.
Nevertheless, it loses the equivalence to convolutions, which negatively impacts its efficiency.

\subsubsection{\textbf{Hardware-aware Computation}}
The selection mechanism is crafted to surpass the limitations of linear time-invariant models.
Still, it challenges efficient training: SSMs' convolutional kernels become input-dependent, resulting in the inability to perform parallel computations.
To tackle the problem, Mamba utilizes two computation techniques, i.e., \emph{Parallel Associative Scan} (also called Parallel Prefix-Sum)~\cite{harris2007parallel} and \emph{Memory Recomputation}.
First, the Parallel Associative Scan leverages the property of linear associative computation and the parallelism of modern accelerators (GPU and TPU) to perform the calculation of selective SSMs in a memory-efficient manner.
More specifically, the parallel associative scan reduces the computation complexity of model training from $\mathbf{O}(N^2d)$ to $\mathbf{O}(N/t)$. 
At its core, the scan revolves around constructing a balanced binary tree on the given input and sweeping it to and from the root.
In other words, the parallel associative scan begins by traversing from the leaves to the root (i.e., Sweep-Up), creating partial sums at the internal nodes of the tree. 
Then, it reverses the traversal, moving from the root back up the tree to construct the whole scan using the partial sums (i.e., Sweep-Down).

On the other hand, Mamba leverages the traditional approach of recomputation to diminish the overall memory demand for training selective SSM layers.
In particular, Mamba abstains from storing intermediate states of size ($B$, $L$, $D$, $N$) during the forward pass of the Parallel Associative Scan to prevent memory expansion.
Instead, it recomputes those intermediate states in the backward pass for gradient computation.
By doing so, recomputation sidesteps the necessity of reading $O(BLND)$ elements between GPU memory cells.
In addition to optimizing the memory needs of the scan operation, Mamba-1 extends its use of recomputation to enhance the efficiency of the entire SSM layer.
This optimization encompasses projections, convolutions, and activations, which typically demand significant memory resources but can be rapidly recomputed.

\subsection{Mamba-2: State Space Duality}
Transformers, which have played a crucial role in the success of deep learning for various areas, have inspired the development of various techniques, such as Parameter-efficient Fine-tuning~\cite{kojima2022large}, Catastrophic Forgetting Mitigation~\cite{korbak2022controlling}, and Model Quantization~\cite{xiao2023smoothquant}, aimed at improving model performance from diverse perspectives. 
To enable state space models to access and benefit from the valuable techniques initially developed for Transformers, Mamba-2~\cite{mamba2} has introduced a comprehensive framework called Structured State-Space Duality (SSD), which establishes theoretical connections between SSMs and different forms of attention.
Formally, 
\begin{align}
    \mathbf{y} = \textbf{SSD}(\mathbf{A}, \mathbf{B}, \mathbf{C})(\mathbf{x}) = \mathbf{M} \mathbf{x},
\end{align}
where $\mathbf{M}$ denotes the matrix form of SSMs that uses the sequentially semi-separable representation, and $\mathbf{M}_{ji} = \mathbf{C}^\text{T}_j \mathbf{A}_{j:i} \mathbf{B}_i$.
Notably, $\mathbf{C}_j$ and $\mathbf{B}_i$ represent the selective space state matrices associated with input tokens $\mathbf{x}_j$ and $\mathbf{x}_i$, respectively.
$\mathbf{A}_{j:i}$ denotes the selective matrix of hidden states corresponding to the input tokens ranging from $j$ to $i$.
In essence, SSD demonstrates that both the attention mechanism used by Transformers and the linear time-variant system employed in SSM can be seen as semi-separable matrix transformations.
Furthermore, \citet{mamba2} also prove that the selective SSM is equivalent to a structured linear attention mechanism implemented with a semi-separable masking matrix.

{
It is notable that SSD requires significantly fewer parameters (i.e., $B \times L$) to represent all $\mathbf{A}$ matrices, compared to conventional SSMs ($B \times L \times N \times N$) and Mamba1 ($B \times L \times N$).
This reduction is due to the specific constraints imposed on the state matrices $\mathbf{A}$: conventional SSMs use full matrices, Mamba1 restricts $\mathbf{A}$ to diagonal matrices, and Mamba2 further simplifies it to a scalar multiple of the identity matrix (i.e., $\alpha \mathbf{I}$).
}
Moreover, based on SSD, Mamba-2 has devised a more hardware-efficient computation through a block decomposition matrix multiplication algorithm.
% , as depicted in Figure~\ref{fig:ssd}
Specifically, by viewing state space models as semi-separable matrices through the matrix transformation, Mamba-2 decomposes the computation into matrix blocks, in which diagonal blocks represent intra-chunk computations. In contrast, the off-diagonal blocks represent inter-chunk computations factored through the SSM’s hidden state.
This approach enables Mamba-2 to achieve a 2-8$\times$ faster training process than Mamba-1's parallel associative scan while remaining competitive with Transformers.

\begin{figure}[t]
\centering
{\includegraphics[width=0.7\linewidth]{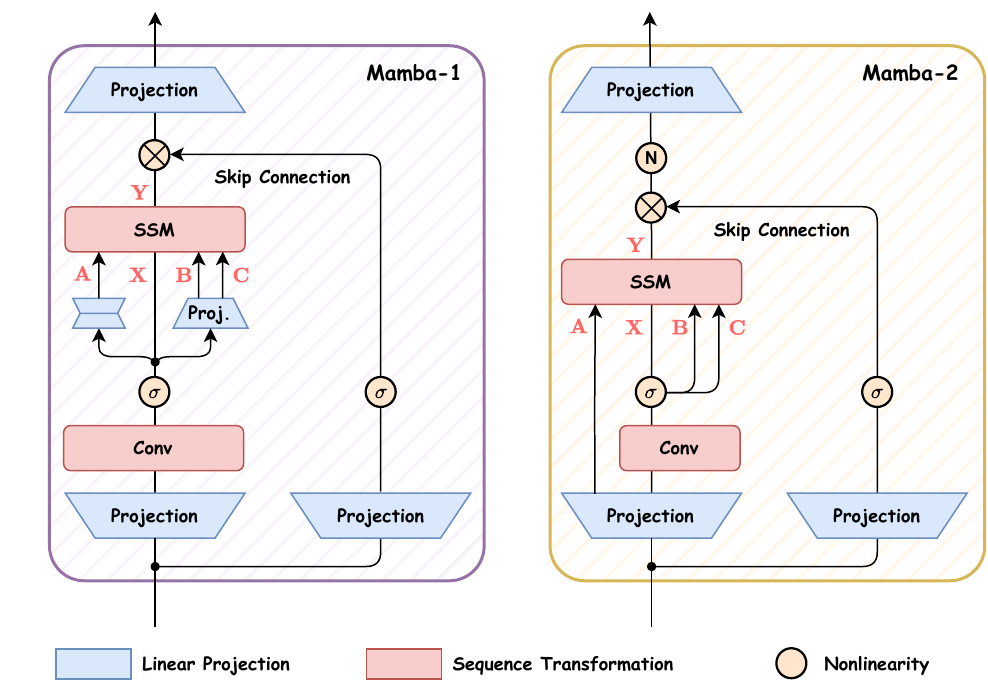}}
\vskip -0.1in
\caption{The block architectures of Mamba-1 and Mamba-2. }
\label{fig:block}
\end{figure}

\subsection{Mamba Block}
In this subsection, we provide a summary of the block design for Mamba-1 and Mamba-2.
Figure~\ref{fig:block} illustrates the comparison of these two architectures.
Mamba-1 is motivated by an SSM-centric point of view where the selective SSM layer is tasked with conducting a map from input sequences $\mathbf{X}$ to $\mathbf{Y}$.
In this design, the linear projections of ($\mathbf{A}$, $\mathbf{B}$, $\mathbf{C}$) are applied after the initial linear projection that creates $\mathbf{X}$.
The input tokens and state matrices are then passed through the selective SSM cell, utilizing the parallel associative scan, to produce the output $\mathbf{Y}$.
After that, Mamba-1 employs a skip connection to encourage feature reuse and alleviate the degradation problem often occurring during the model training process.
Finally, the Mamba model is constructed by stacking this block, interleaved with standard normalization and residual connections. 

As for Mamba-2, it introduces the SSD layer aiming to create a map from $[\mathbf{X}$, $\mathbf{A}$, $\mathbf{B}$, $\mathbf{C}]$ to $\mathbf{Y}$.
This is achieved by simultaneously processing $[\mathbf{X}$, $\mathbf{A}$, $\mathbf{B}$, $\mathbf{C}]$ with a single projection at the beginning of the block, similar to how standard attention architectures generate the $\mathbf{Q}$, $\mathbf{K}$, $\mathbf{V}$ projections in parallel.
In other words, the Mamba-2 block simplifies the Mamba-1 block by removing sequential linear projections.
This enables faster computation of the SSD structure compared to the parallel selective scanning in Mamba-1.
Additionally, a normalization layer is added after the skip connection, aiming to improve training stability.

% Table generated by Excel2LaTeX from sheet 'Sheet1'
\begin{table}[htbp]
  \centering
  \caption{{Representative Open-Access Foundation Models Utilizing Mamba Architecture.}}
  \vskip -0.1in
    \begin{tabular}{cccrc}
    \toprule
    \textbf{Name} & \textbf{Modality} & \textbf{Affiliations} & \multicolumn{1}{c}{\textbf{Sizes}} & \multicolumn{1}{c}{\textbf{Access Link}} \\
    \midrule
    Mamba 1\&2 & Language & Carnegie Mellon University \& Princeton University & 130M-2.8B & 1 \\
    Falcon Mamba 7B & Language & Technology Innovation Institute & 7B    & 2 \\
    Mistral 7B & Language & Mistral AI \& NVIDIA & 7B    & 3 \\
    Jamba & Language & AI21 Lab & 12B/52B & 4 \\
    Vision Mamba & Vision & Huazhong University of Science and Technology & 7M-98M & 5 \\
    VideoMamba & Video & OpenGVLab, Shanghai AI Laboratory & 28M-392M & 6 \\
    Codestral Mamba & Code & Mistral AI & 7B, 22B    & 7 \\
    SMI-SSED & Molecule & IBM Research & 1.3B    & 8 \\    
    \midrule
    \multicolumn{5}{l}{1. \url{https://github.com/state-spaces/mamba}} \\
    \multicolumn{5}{l}{2. \url{https://huggingface.co/tiiuae/falcon-mamba-7b}} \\
    \multicolumn{5}{l}{3. \url{https://huggingface.co/mistralai/Mistral-7B-v0.1}} \\
    \multicolumn{5}{l}{4. \url{https://huggingface.co/ai21labs/Jamba-v0.1}} \\    
    \multicolumn{5}{l}{5. \url{https://huggingface.co/hustvl/Vim-base-midclstok}} \\
    \multicolumn{5}{l}{6. \url{https://huggingface.co/OpenGVLab/VideoMamba}} \\
    \multicolumn{5}{l}{7. \url{https://mistral.ai/news/codestral-mamba}} \\
    \multicolumn{5}{l}{8. \url{https://huggingface.co/ibm-research/materials.smi_ssed}} \\
    \bottomrule
    \end{tabular}%
  \label{tab:model}%
\end{table}%

\section{Advancements in Mamba Models}
\label{se:module}
State Space Models and Mamba have been recently explored and have become a promising alternative as the foundational model backbone.
As shown in Table~\ref{tab:model}, large-scale Mamba-based models have not only thrived within academic research but have also made significant strides in industry, such as \emph{Falcon Mamba 7B} and \emph{Mistral 7B}, demonstrating their efficacy through successful training on GPUs.
Despite that, the Mamba architecture still encounters challenges, such as memory loss, generalization to diverse tasks, and inferior capability to capture complex patterns compared to Transformer-based language models.
To overcome these challenges, plenty of efforts have been made to improve the Mamba architecture.
Existing research studies primarily concentrate on modifying the \emph{block design}, \emph{scanning mode}, and \emph{memory management} aspects. 
This section will introduce several vital techniques from these three aspects, and a summary of related studies is presented in Table~\ref{tab:module}.

\begin{table*}[htbp]
  \centering
  \caption{Summary of Existing Studies on Improving the Mamba Model.}
  \vskip -0.1in
  \scalebox{0.9}{
    \begin{tabular}{c|c|c|l}
    \toprule
    \textbf{Modules} & \textbf{Methods} & \textbf{Classes} & \textbf{Representative References} \\
        \midrule
        \multirow{11}[6]{*}{Block} & \multirow{5}[2]{*}{Integration} & Transformer & \cite{lieber2024jamba,xu2024integrating,pilault2024block,hatamizadeh2024mambavision,pitorro2024effective,gao2024matten,park2024can} \\
          &       & Convolutional Neural Network (CNN)   & \cite{li2024mambadfuse,wang2024weak,yue2024medmamba,yang2024hsimamba,gong2024nnmamba,sheng2024dualmamba,yuan2024st} \\
          &       & Graph Neural Network (GNN)  & \cite{liu2024mamba4rec,li2024stg,behrouz2024graph,wang2024graph,yang2024uncovering} \\
          &       & Recurrent Neural Network (RNN)   &  \cite{tang2024vmrnn,dolga2024rotrnn,huang2024recurrent} \\
          &       & Spiking Neural Network (SNN)   & \cite{li2024spikemba,bal2024rethinking} \\
\cmidrule{2-4}          & \multirow{3}[2]{*}{Substitution} & U-Net  & \cite{sepehri2024serpent,shi2024vmambair,wang2024mamba,wang2024large,liu2024swin,ruan2024vm,liao2024lightm,ma2024semi,sanjid2024integrating,deng2024cu,ji2024self,hosseini2024sum} \\
          &       & Diffusion Models & \cite{oshima2024ssm,fu2024md,fei2024scalable,ye2024p,wang2024weak} \\
          &       & Others & \cite{chen2024res,li2024spmamba} \\
\cmidrule{2-4}          & \multirow{3}[1]{*}{Modification} & Mix-of-Expert   & \cite{lieber2024jamba,anthony2024blackmamba} \\
          &       & K-way/Parallel Structure & \cite{wu2024ultralight,wan2024sigma,zou2024rhythmmamba,huang2024localmamba,lin2024mtmamba} \\
            &       & Register & \cite{wang2024mamba,yang2024cmvim} \\

    \midrule
    \multirow{6}[6]{*}{Scan} & \multirow{4}[1]{*}{Flatten} & Bidirectional Scan & \cite{zhu2024vision,jiang2024dual,li2024spmamba,li2024harmamba,qu2024ssd4rec}  \\
              &  & Sweeping Scan & \cite{liu2024vmamba,wang2024vmambamorph,yue2024medmamba} \\
          &  & Continuous Scan & \cite{yang2024plainmamba,hu2024zigma,he2024mambaad} \\
           &  & Efficient Scan & \cite{pei2024efficientvmamba,xie2024fusionmamba} \\

\cmidrule{2-4}          & \multirow{3}[1]{*}{Stereo} & Hierarchical Scan & \cite{chen2024mim,wang2024large,bhirangi2024hierarchical,chen2024survmamba,han2024mamba3d,shi2024multi} \\
          &       & Spatiotemporal Scan & \cite{li2024videomamba,chen2024changemamba,yao2024spectralmamba,yang2024vivim} \\
          & & Hybrid Scan & \cite{behrouz2024mambamixer,shi2024vmambair,gong2024nnmamba,he2024pan,dong2024fusion,deng2024cu} \\
    \midrule
Memory & \multicolumn{3}{c}{Initialization~\cite{ezoe2024learning}, Compression~\cite{long2024dgmamba,nawrot2024dynamic}, Connection~\cite{he2024densemamba,ren2024samba}, State-Tracking~\cite{merrill2024illusion} } \\
    \midrule
Others & \multicolumn{3}{c}{Autoregressive Pretraining~\cite{ren2024autoregressive}, Explainability~\cite{jafari2024mambalrp}} \\
    \bottomrule
    \end{tabular}%
    }
  \label{tab:module}%
\end{table*}%

\subsection{Block Design}
\begin{figure}[t]
\centering
{\includegraphics[width=0.95\linewidth]{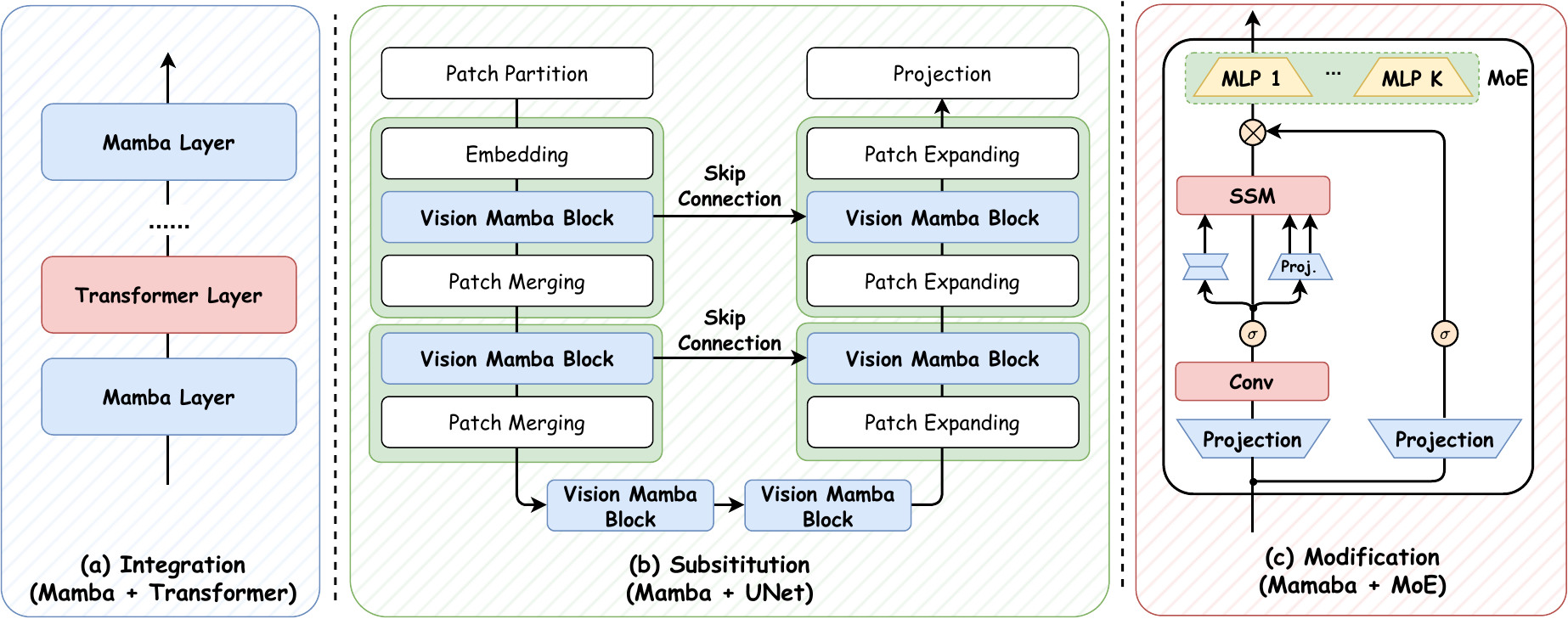}}
\caption{Representative examples of improved Mamba models based on the perspective of block design: 
(a) Integration methods combine orthogonal architectural designs (e.g., Transformer) with Mamba, leading to enhanced model performance and increased throughput, all while upholding a manageable memory footprint~\cite{xu2024integrating,pilault2024block}; 
(b) Substitution methods enhance the learning capabilities of standard learning frameworks (e.g., UNet) by integrating the Mamba block as a substitute for their primary layers~\cite{yue2024medmamba,liao2024lightm};
(c) Modification methods employ cutting-edge techniques, such as Mix-of-Expert (MoE), to refine the Mamba block~\cite{lieber2024jamba}.
}
\label{fig:module}
\end{figure}

The design and structure of the Mamba block have a significant impact on the overall performance of Mamba models, making it an emerging research focus. 
As illustrated in Figure~\ref{fig:module}, based on different approaches to constructing new Mamba blocks,   existing research can be categorized into three categories: a) \textbf{Integration} methods aim to integrate the Mamba block with other well-known models, so as to strike a balance between effectiveness and efficiency; b) \textbf{Substitution} methods attempt to utilize Mamba block as a substitution for main layers in advanced model frameworks; and c) \textbf{Modification} methods focus on modifying the components within the classical Mamba block.
Accordingly, we will present a detailed review of these methods in the following subsections.

\subsubsection{\textbf{Integration}}
Given Mamba's exceptional ability to capture long-term dynamics, it has been extensively integrated with other models, leveraging its strengths to deliver a robust framework tailored to specific scenarios.
The integration specifically encompasses advanced models like Transformers, Graph Neural Networks (GNNs), Recurrent Neural Networks (RNNs), Convolutional Neural Networks (CNNs), and Spiking Neural Networks (SNNs). 
Specific examples are described below.

\begin{itemize}[leftmargin=*]
    \item \textbf{Transformer}-based models have exhibited remarkable performance in numerous tasks, but their quadratic computational complexity still hampers them during the inference process~\cite{gu2021efficiently}.
    In the pursuit of efficient generation, some researchers have proposed incorporating Mamba blocks with Transformer-based models~\cite{pilault2024block,hatamizadeh2024mambavision,pitorro2024effective,gao2024matten,park2024can}.
    For example, Jamba~\cite{lieber2024jamba} combines blocks of Transformer and Mamba layers to tackle long-content Natural Language Processing tasks, capitalizing on the advantages of both model families.
    The Attention-Mamba hybrid model demonstrated superior performance compared to the standalone Transformer and Mamba models, achieving better throughput than the vanilla Transformer model.
    Mambaformer~\cite{xu2024integrating} utilizes the hybrid framework to forecast multiple time series, which internally combines Mamba blocks and Transformer layers for long- and short-range dependencies, respectively.
    Due to the integration of Mamba and Transformer, Mambaformer outperforms Transformer-based predictors in long-short range time series forecasting.

    \item \textbf{GNN} has demonstrated promising potential in capturing neighboring relationships through message-passing mechanisms, where information is propagated over a connection graph through stacked layers.
    Nonetheless, these models face a significant limitation known as over-smoothing~\cite{chen2020measuring}, particularly when attempting to capture high-order adjacency signals.
    To tackle such a challenge, Mamba has been employed for graph representation learning~\cite{liu2024mamba4rec,li2024stg,yang2024uncovering,wang2024graph}.
    For example, Graph Mamba~\cite{behrouz2024graph} reformulates graph-structured data into sequential tokens in a particular order and leverages a selective SSM layer within the Mamba block to construct a novel Graph Mamba Network (GMN) architecture, which achieves superior graph representation learning capabilities, particularly in the datasets that require high-order dependencies between nodes.

    \item  \textbf{RNN}-based models have yielded outstanding results in capturing temporal dynamics.
    Nevertheless, RNNs still face significant challenges, including time-consuming recurrent training and limitations in memory capacity for hidden states.
    Inspired by the emergence of recent Mamba-based architectures, some researchers have developed a fusion of Mamba blocks and RNNs.
    For instance, VMRNN~\cite{tang2024vmrnn} achieves state-of-the-art performance in spatio-temporal forecasting while minimizing floating-point operations (FLOPs) compared to recurrent-based and recurrent-free methods.
    It accomplishes this by introducing a novel recurrent unit that combines Mamba blocks with Long Short-Term Memory (LSTM).

    \item \textbf{CNN}-based methods are constrained by local receptive fields, resulting in suboptimal performance capturing global and long-range semantics~\cite{gu2023mamba}.
    Known for the superior capability of state space models to learn long-range patterns, some studies~\cite{wang2024weak,li2024mambadfuse,yang2024hsimamba} have explored the potential of utilizing Mamba blocks to enhance CNN-based models, especially in the field of computer vision.
    For instance, MedMamba~\cite{yue2024medmamba} and nnMamba~\cite{gong2024nnmamba} showcase how the integration of visual Mamba blocks improves the performance of CNNs in image analysis tasks.
    
    \item \textbf{SNN} has been recently proposed as a promising network architecture inspired by the behavior of biological neurons in the brain: transmitting knowledge between neurons through discrete spikes.
    One of the key advantages of SNNs lies in their potential for low-power implementation, as they can exploit the sparse and event-driven nature of neural activity.
    Motivated by the energy-efficient implementation of SNNs and SSMs' superior long-range learning capabilities, pioneering studies have delved into integrating these two methods.
    For example, SpikeMba~\cite{li2024spikemba} combines them to handle confidence bias towards prominent objects and to capture enduring dependencies within video sequences.
    Through extensive evaluations, the authors claim that integrating these two models improves the effectiveness of temporal video grounding tasks, precisely moment retrieval and highlight detection.

\end{itemize}

\subsubsection{\textbf{Substitution}}
Inspired by the outstanding capabilities of the selective SSM in efficient computation and long sequence learning, the adoption of Mamba modules to replace critical components in classical modeling frameworks such as U-Net~\cite{ronneberger2015u} and Diffusion Model~\cite{ho2020denoising} has attracted a lot of attention.
By introducing the selective SSM layer, these methods achieve long-range learning and efficient computation for their specific tasks.
Below, we demonstrate instances of substitution using the Mamba module, specifically for advanced frameworks such as U-Net and Diffusion models.

\begin{itemize}[leftmargin=*]
    \item \textbf{U-Net}.  Many efforts~\cite{shi2024vmambair,wang2024mamba,wang2024large,liao2024lightm} have been made to synergize U-Net with Mamba’s capability in capturing intricate and broad semantics so as to advance model performance in computer vision tasks.
    For example, Mamba-UNet~\cite{wang2024mamba} utilizes Visual Mamba blocks exclusively to construct a U-Net-like model (i.e., an encoder-decoder model infused with skip connections) for medical image segmentation.
    Their evaluation demonstrates that Mamba-UNet surpasses several U-Net variations, which can be attributed to the efficacy and efficiency of Mamba blocks in handling long-range patch sequences.

    \item \textbf{Diffusion Model}. Some endeavors~\cite{fu2024md,fei2024scalable,oshima2024ssm} have been undertaken to build a novel type of diffusion model, Diffusion State Space Model (DiS), which replaces the typical backbone (e.g., CNNs, Attentions, U-Nets) with a state space backbone.
    Given the remarkable efficiency and efficacy of Mamba blocks in accommodating long-range dependencies, DiS is distinguished by generating longer sequences using diffusion models~\cite{fei2024scalable}.
    For example, \citet{oshima2024ssm} propose a Mamba-based diffusion model that substantially decreases memory consumption for long video sequences, while still maintaining competitive performance metrics when compared to Transformer-based models.
    Moreover, MD-Dose~\cite{fu2024md} and P-Mamba~\cite{ye2024p} construct noise predictors using Mamba blocks in the backward process of diffusion models, ultimately generating specific targets for medical image processing.

    \item \textbf{Others}. Besides the U-Net and Diffusion Models, there are a few substitutions.
    For example, Res-VMamba~\cite{chen2024res} adopts Visual Mamba blocks in a residual learning framework for food category classification.
    Furthermore, SPMamba~\cite{li2024spmamba} adopts the TF-GridNet~\cite{wang2023tf}, a recently developed time-frequency model, as its base architecture, followed by replacing the Transformer components with bidirectional Mamba blocks.
    This adaptation enables the model to efficiently encompass a wider scope of contextual information for the task of speech separation.
\end{itemize}

\subsubsection{\textbf{Modification}}
Apart from integration and substitution methods that directly employ the Mamba block, some other efforts have been made to modify the Mamba block with the aim of enhancing its performance in different scenarios.
For example, Jamba~\cite{lieber2024jamba} borrows the conception of Mix-of-Experts (MoE)~\cite{jacobs1991adaptive,fedus2022switch} to enable their hybrid (Transformer-Mamba) decoder-only model to be pretrained with far less compute and allow flexible objective-specific configurations.
Notably, the Jamba model (56B available parameters, 12B active parameters, 4GB KV cache) requires a 32x smaller KV cache compared to a representative Transformer-based language model, LLaMA-2-7B (6.7B available parameters, 12B active parameters, 128GB KV cache), while providing more extensive available and active parameters.
This allows Jamba to swallow a context length of 140K on a single A100 GPU (80GB), seven times the length supported by LLaMA-2-70B. 
In addition to MoE, some studies propose modifying the SSM layer into a K-way structure, which involves processing model inputs using parallel SSM cells, allowing for capturing information and knowledge from multiple perspectives. 
For example, Sigma~\cite{wan2024sigma} develops a novel Mamba-based visual encoder that handles multimodal inputs by utilizing parallel SSM layers.
UltraLight VM-UNet~\cite{wu2024ultralight} proposes a vision Mamba layer with parallel SSM cells that process deep features in different channels.
To recap, by implementing such modifications (i.e., K-way, MoE), these Mamba-based models gain enhanced learning capabilities, particularly in processing multimodal inputs and fast adapting to multiscale tasks. 
In addition, a pioneering study, Mamba$^{\circledR}$, has introduced a novel approach that suggests incorporating registers evenly within the visual input tokens before passing the inputs through the SSM layers.
This modification aims to enhance the representation of the sequence direction of image patches, thereby enabling the unidirectional inference paradigm of the Mamba block to be applicable to visual tasks.
Despite these successes, the exploration of modifying Mamba blocks remains a promising yet under-explored area.

\subsection{Scanning Mode}
\begin{figure}[htbp]
\centering
{\includegraphics[width=0.9\linewidth]{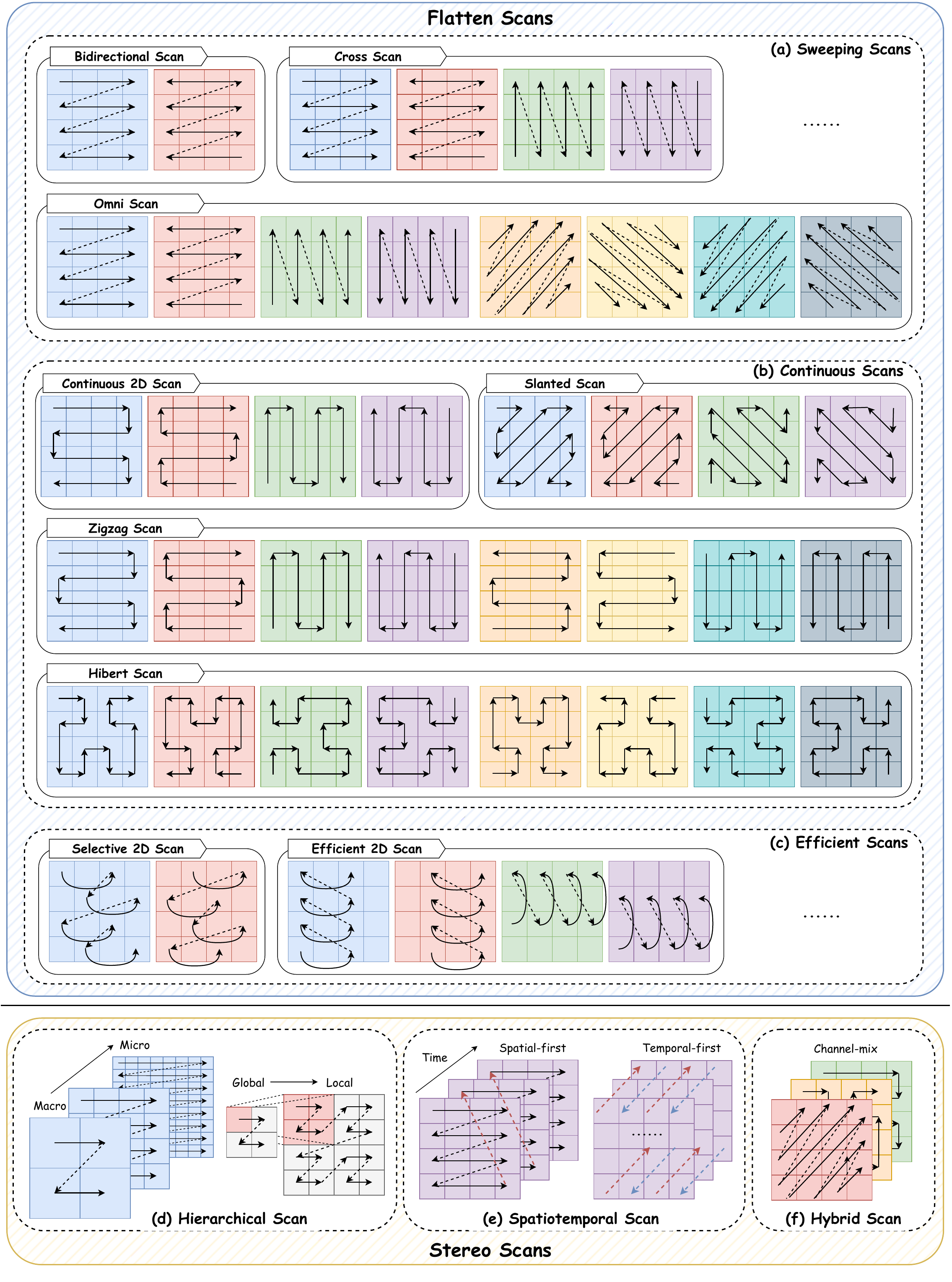}}
\vskip -0.2in
\caption{Recently developed scanning methods in Mamba-based models: Flatten Scans (a-c) involve flattening the model input into token sequences and scanning them accordingly from different directions, and Stereo Scans (d-e) process inputs from additional perspectives, capturing a broader spectrum of knowledge compared to Flatten Scan methods.}
\label{fig:scan}
\end{figure}

The parallel associative scan operation serves as a crucial component within the Mamba model, which aims to address the computation problem caused by the selection mechanism, accelerate the training process, and reduce memory requirements.
It achieves this by leveraging the linear property of time-varying SSMs to design kernel fusion and re-computation at the hardware level.
However, Mamba's uni-directional sequence modeling paradigm hinders a comprehensive learning process for various data, such as images and videos.
To mitigate this issue, several studies have focused on designing efficient scanning methods to enhance model performance and facilitate the training process of Mamba models.
As shown in Figure~\ref{fig:scan}, existing studies that concentrate on developing the scanning mode techniques can be categorized into two classes: 1) \textbf{Flatten Scan} approaches process model inputs from a flat perspective of token sequence; and 2) \textbf{Stereo Scan} methods scan model inputs across dimensions, channels, or scales.

\subsubsection{\textbf{Flattening Scan}}
Flattening scan refers to the process of flattening the model input into token sequences and scanning them accordingly from different directions.
This type of scanning is commonly employed for both one-dimensional (e.g., time series) and two-dimensional (e.g., image) data.
In this section, we further categorize it into four classes, namely \textbf{Bidirectional Scan}, \textbf{Sweeping Scan}, \textbf{Continuous Scan}, and \textbf{Efficient Scan}.

\begin{itemize}[leftmargin=*]
    \item \textbf{Bidirectional Scan}. Borrowing the conception of bidirectional recurrent neural networks (Bi-RNNs)~\cite{schuster1997bidirectional}, Visual Mamba~\cite{zhu2024vision} introduces a scanning method for visual data, called Bidirectional Scan (Bi-Scan), which involves processing input tokens using simultaneous forward and backward SSMs, thus enhancing the model capacity for spatially-aware processing.
    Recently, a number of studies have leveraged the Bi-Scan method to facilitate the learning abilities of their Mamba-based models~\cite{li2024harmamba}.
    For example, DPMamba~\cite{jiang2024dual} and SPMamba~\cite{li2024spmamba} have both leveraged a pair of dual-path (forward and backward) selective SSMs to model the dependency of speech signals, enabling a bidirectional knowledge processing for speech separation.
    Such notable successes can be attributed to the effectiveness of Bi-Scan and its ease of deployment.

    \item \textbf{Sweeping Scan.} As illustrated in Figure~\ref{fig:scan}, the Sweeping Scan technique processes the model inputs in a specific direction, analogous to a cleaner meticulously sweeping a floor~\cite{yue2024medmamba,wang2024vmambamorph}.
    For instance, Cross-Scan~\cite{liu2024vmamba} entails dividing the input image into patches and subsequently flattening it along four distinct paths, which is regarded as a fusion of two bidirectional scans.
    By adopting these complementary traversal paths, Cross Scan enables each patch in the image to efficiently integrate information from its neighbors in different directions, thereby facilitating the establishment of informative, receptive fields.
    Omni-Scan~\cite{shi2024vmambair,zhao2024rs} incorporates the modeling of image information flows from multiple directions, e.g., 2 (forward and backward) $\times$ 4 (left-right, top-bottom, top right-bottom left, top left-bottom right).
    Such a strategy augments the global modeling capability of contextual information in various directions, enabling the extraction of comprehensive global spatial features.

    \item \textbf{Continuous Scan.} To ensure the continuity of input sequences, Continuous Scan techniques scan the adjacent tokens between columns or rows~\cite{he2024mambaad}, as shown in Figure~\ref{fig:scan}.
    For example, in order to better cope with 2D spatial inputs, PlainMamba~\cite{yang2024plainmamba} introduced a continuous scanning approach, known as Continuous Scan, which scans the adjacent tokens between columns (or rows), instead of traveling to the opposite tokens in Cross Scan.
    Moreover, Hilbert Scan~\cite{he2024mambaad} travels a sinuous path based on the Hilbert matrix.
    Based on their evaluation results, it can be inferred that enhancing the semantic continuity of input tokens leads to superior performance in various visual recognition tasks for Mamba-based models.

    \item \textbf{Efficient Scan}. In contrast to the aforementioned scanning methods, which focus on achieving a more comprehensive input modeling, efficient scanning methods aim to accelerate the training and inference process.
    Generally, the efficient scan separates the given input into several parts and processes them in parallel, thus reducing computational time.
    For example, Efficient-2D Scan~\cite{pei2024efficientvmamba} proceeds images by skipping patches, thus reducing computational demands by a factor of four while preserving global feature maps.
    Moreover, \citet{gao2024mamba} introduce an effective bi-directional subspace scanning scheme within their Mamba framework.
    This scheme is designed to capture long-term spatial-angular correspondences efficiently for 4D light field super-resolution tasks.
    Specifically, it decomposes the patch sequences into two parts and processes them through two bi-directional scanning schemes.
    By doing so, the scanning method lowers the input length and addresses the long-term memory issues without sacrificing the complete 4D global information.
    
\end{itemize}

\subsubsection{\textbf{Stereo Scan}}
By modeling inputs from additional perspectives, stereo-scan methods excel in capturing a broader spectrum of knowledge during the scanning process when compared to flattened scan methods.
This enhanced capability allows for a more thorough comprehension of model inputs. 
To be specific, these methods can be classified into three primary categories: \textbf{Hierarchical Scan}, \textbf{Spatiotemporal Scan}, and \textbf{Hybrid Scan}.
The Hierarchical Scan processes the input from different levels, while the Spatiotemporal Scan considers input patterns from both temporal and spatial perspectives.
Additionally, Hybrid Scan combines multiple scanning methods to leverage the benefits of different scan techniques.

\begin{itemize}[leftmargin=*]
    \item \textbf{Hierarchical Scan} methods involve employing different kernel sizes of scanning to capture the semantic knowledge from global to local or from macro to micro perspectives~\cite{wang2024large,chen2024survmamba,han2024mamba3d,shi2024multi}.
    For example, a Mamba-in-Mamba hierarchical encoder is proposed by \cite{chen2024mim} for infrared small target detection, combining inner and outer selective SSM blocks.
    The inner one is specifically tailored to capture the interplay among visual patches for local pattern extraction.
    Conversely, the outer block is designed to characterize the relationship between visual sentences to capture global features.
    HiSS~\cite{bhirangi2024hierarchical} divides an input sequence into chunks and models the chunk features hierarchically for continuous sequential prediction.
    The chunks are first processed by a low-level SSM cell, and the processed features are mapped into an output sequence by a high-level SSM block.
    
    \item \textbf{Spatiotemporal Scan}. Driven by the prevalence of dynamic systems in the real world, there has been a growing interest in spatiotemporal scanning methods to enhance the performance of Mamba block~\cite{yao2024spectralmamba, yang2024vivim}.
    For instance, VideoMamba~\cite{li2024videomamba} expands the original 2D scan for images into two 3D scans: spatial-first scanning and temporal-first scanning.
    Combining these two scanning approaches, VideoMamba demonstrates exceptional efficiency in handling long, high-resolution videos.
    Additionally, ChangeMamba~\cite{chen2024changemamba} integrates three spatiotemporal scanning mechanisms (sequential modeling, cross modeling, and parallel modeling) to enable contextual information interaction among multi-temporal features for remote sensing change detection.
    
    \item \textbf{Hybrid Scan}. In the pursuit of comprehensive feature modeling, many efforts have focused on combining the advantages of different scanning methods~\cite{zhen2024freqmamba,shi2024vmambair,gong2024nnmamba,dong2024fusion,deng2024cu}, so-called Hybrid Scan.
    For example, Mambamixer~\cite{behrouz2024mambamixer} presents Switch of Scan that dynamically employs a set of image scanning methods, namely Cross-Scan, Zigzag Scan, and Local Scan, to traverse image patches.
    Mambamixer also introduces a dual selection mechanism to mix information across tokens and channels.
    By doing so, they show competitive performance with other vision models.
    Pan-Mamba~\cite{he2024pan} introduces two scanning methods built upon the Mamba architecture: channel swapping scan and cross-modal scan.
    By incorporating these two scanning approaches, Pan-Mamba enhances its capabilities in efficient cross-modal information exchange and fusion for image pan-sharpening.
\end{itemize}

\subsection{Memory Management}
Like RNNs, the memory of hidden states within state space models effectively stores information from previous steps, thereby playing a crucial role in SSM's overall functionality.
While Mamba has introduced the HiPPO-based method for memory initialization~\cite{gu2023mamba}, challenges still exist in the memory management of the SSM cell, including transferring hidden information between layers and achieving lossless memory compression.
To this end, a handful of pioneering studies have proposed different solutions~\cite{wang2024memorymamba}.
For example, \citet{ezoe2024learning} have attempted to refine the initialization process of selective SSMs by using a balanced truncation method during model retraining.
Moreover, DGMamba~\cite{long2024dgmamba} introduces a Hidden State Suppressing method to bolster the domain generalization capabilities of the hidden states within State Space Models.
This method works to alleviate the negative effects stemming from these hidden states, thereby narrowing the gap between hidden states across different domains.
On a similar note, DenseMamba~\cite{he2024densemamba} has put forth a dense connection method to enhance the propagation of hidden information between layers in SSMs.
This strategy aims to mitigate memory degradation and preserve detailed information for output generation by selectively integrating hidden states from shallower layers into deeper ones.
Moreover, \citet{merrill2024illusion} delve into the concept of the illusion of "State" in SSMs through extensive experiments, revealing their constraints in addressing practical state-tracking challenges like monitoring chess moves.
Expanding on this, the researchers enhance SSMs with a subtle modification: by introducing input-dependent transition matrices, enabling effective state tracking and permutation composition.

\section{Adapting Mamba to Diverse Data}
\label{se:data}
\begin{figure}[htbp]
\centering
{\includegraphics[width=0.95\linewidth]{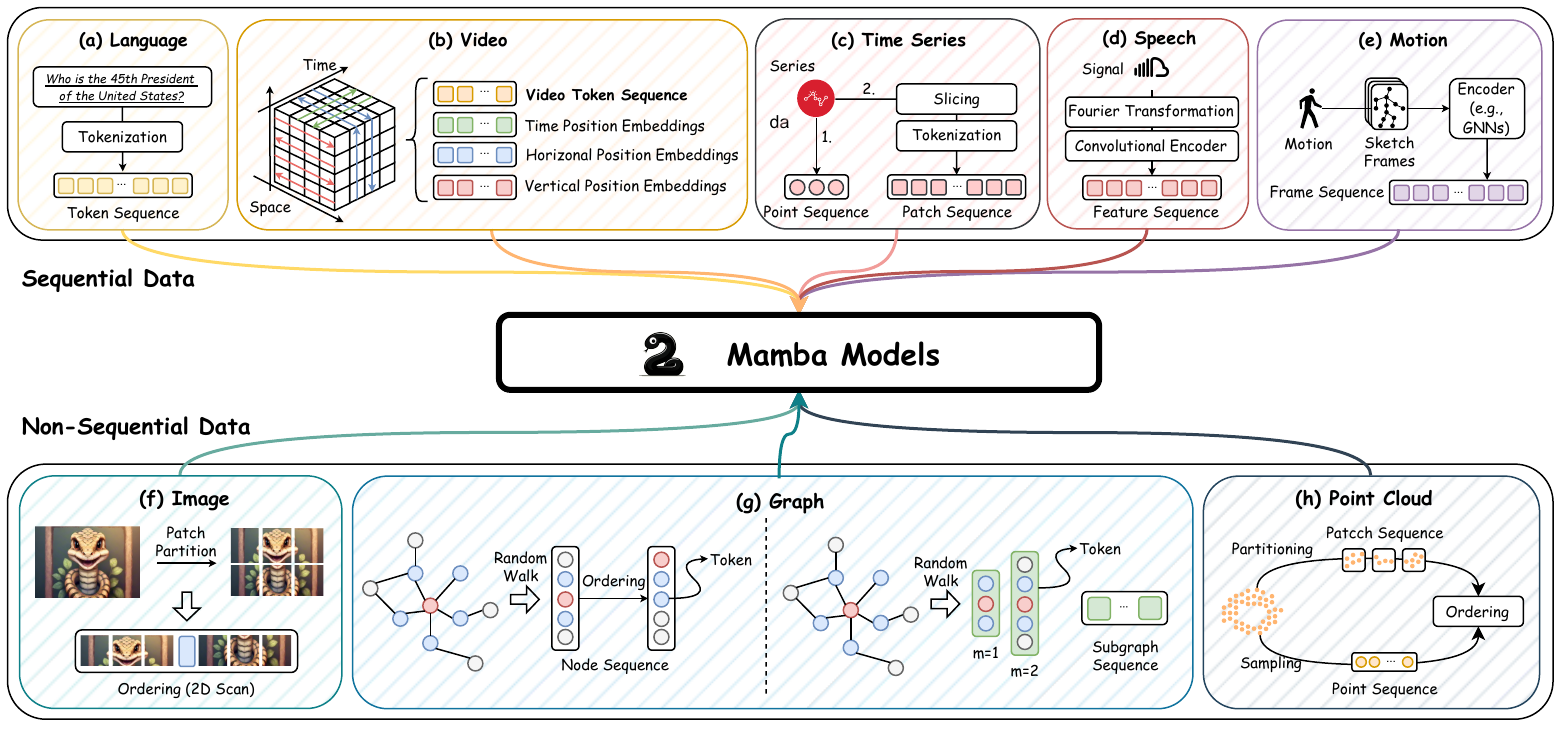}}
\vskip -0.1in
\caption{Representative strategies exist for adapting Mamba to diverse types of data. (a-e) The Mamba architecture, imbued with essential characteristics of recurrent models, serves as an ideal foundational model for handling sequences like language, time series, and speech. (f-h) A common approach to address non-sequential data involves segmenting or sampling the data into discrete tokens and organizing them into sequences following a defined rule. Additionally, Mamba exhibits the capability to process multimodal data by either concatenating their respective sequences or projections.}
\label{fig:data}
\end{figure}

The Mamba architecture represents an extension of selective state space models, which possess fundamental properties of recurrent models that make it well-suited as a general foundation model operating on sequences like text, time series, speech, and more.
Meanwhile, recent pioneering studies have extended the utilization of the Mamba architecture beyond sequential data, encompassing domains such as images and graphs, as depicted in Figure~\ref{fig:data}.
These studies aim to harness Mamba's remarkable capabilities in capturing long-range dependencies while leveraging its efficiency in learning and inference processes. 
In this section, we therefore aim to investigate the emerging techniques that adapt Mamba to various types of data.
A summary of related studies is illustrated in Table~\ref{tab:data-mamba}.

\begin{table*}[htbp]
  \centering
  \caption{Summary of Mamba-associated research in different types of data.}
  \scalebox{0.9}{
    \begin{tabular}{llll}
    \toprule
    Category & Data  & Typical Tasks & Representative References \\
    \midrule
    Sequential Data & Language &    Long-Context Language Modelling   & \cite{shen2021efficient,poli2023hyena,gu2021efficiently,behrouz2024mambamixer,bhirangi2024hierarchical,nawrot2024dynamic,he2024densemamba,lieber2024jamba,anthony2024blackmamba} \\
           & Video &    Long Video Generation   & \cite{yang2024vivim,li2024spikemba,oshima2024ssm,zou2024rhythmmamba,arnab2021vivit,liu2022video,li2024videomamba} \\
         & Time Series &   Long-Term Forecasting   & \cite{xu2024integrating,liang2024bi,ahamed2024tscmamba,zhou2021informer,ahamed2024timemachine,sanjid2024integrating,yuan2024st} \\
          & Speech &    Speech Separation   & \cite{abdel2014convolutional,chen2024rawbmamba,li2024spmamba,jiang2024dual} \\
          & Motion &   Continuous Human Motion Understanding    & \cite{wang2024text,zhang2024motion,zeng2024mambamos} \\
          \midrule
    Non-Sequential Data & Image &   High-Resolution Medical Vision    & \cite{yue2024medmamba,chen2024rsmamba,chen2024changemamba,lin2024pixmamba,wang2024mamba,ruan2024vm,zhu2024vision,touvron2021training} \\
          & Graph &   Large Graph Learning    & \cite{fan2019graph,huang2020skipgnn,liu2024mamba4rec,ye2021sparse,wang2024graph,behrouz2024graph,huang2024can} \\
          & Point Cloud &   Efficient 3D Point Cloud Restoration   & \cite{guo2020deep,yu2022point,yi2024mvgamba,han2024mamba3d,zhou20243dmambaipf,liang2024pointmamba,zhang2024point} \\
          \midrule
    Multimodal Data & Vision-Languge & Visual and Linguistic Awareness  & \cite{yang2024shmamba,liu2024robomamba,wu2023multimodal,wang2024text,qiao2024vl} \\
    & Multimodality & Semantic Recognition & \cite{dong2024fusion,wan2024sigma} \\
    \bottomrule
    \end{tabular}%
  }
  \label{tab:data-mamba}%
\end{table*}%

\subsection{Sequential Data}
Sequential data refers to data gathered or organized in a particular order, where the order of the data points holds significance.
To explore the potential of utilizing Mamba as a foundation model for tasks concerning sequential data, we provide a comprehensive review presented in the subsequent sections, which covers various sequential data, including natural language, video, time series, speech, and human motion.

\subsubsection{\textbf{Natural Language}}
As one of the most representative architectures, Mamba performs content-based reasoning while ensuring efficiency, which is considered a promising alternative for the backbone of large language models to address Transformers’ computational inefficiency on long sequences. 
Building upon this insight, many studies have explored the potential of Mamba for various downstream tasks in natural language processing (NLP)~\cite{behrouz2024mambamixer,bhirangi2024hierarchical,nawrot2024dynamic,he2024densemamba}.
For example, MambaByte~\cite{wang2024mambabyte} is proposed to utilize Mamba on byte sequences, aiming to leverage the advantages of Mamba in capturing long-range dependencies for token-free language models.
Their evaluations show that MambaByte avoids the inductive bias of subword tokenization and outperforms state-of-the-art subword Transformers on long-term language modeling tasks.
Besides, Jamba~\cite{lieber2024jamba} and BlackMamba~\cite{anthony2024blackmamba} incorporate the concept of Mix-of-Experts (MoE) to enhance Mamba's performance on language processing by integrating the linear-complexity generation from SSMs with the rapid and economical inference capabilities offered by MoE.

\subsubsection{\textbf{Video}}
The core objective for video understanding and generation lies in learning spatiotemporal representations, which inherently presents two formidable challenges: the large spatiotemporal redundancy within short video clips and the complex spatiotemporal dependencies among long contexts~\cite{arnab2021vivit,liu2022video}.
In the pursuit of addressing both challenges simultaneously, Mamba stands out with its capabilities in distinguishing short-term actions and interpreting long videos~\cite{li2024spikemba,oshima2024ssm,zou2024rhythmmamba,gao2024matten}.
For instance, VideoMamba~\cite{li2024videomamba} first projects the input videos into a set of non-overlapping spatiotemporal patches through 3D convolution, and then utilizes stacked bidirectional Mamba blocks to encode these patches into vectorized representations for downstream tasks like video understanding and generation.
% v1_added
Moreover, Vivim~\cite{yang2024vivim} presents a novel temporal Mamba block to effectively compress extensive spatiotemporal representations into multi-scale sequences for medical video segmentation.
    
\subsubsection{\textbf{Time-Series}}
As typical sequential data, time-series data is ubiquitous in various facets of our lives, including stock market analysis, traffic modeling, and weather forecasting~\cite{zhou2021informer,qu2024physics}.
Motivated by the recent progress on Mamba in modeling long-range sequences, many efforts have been made to investigate its potential for time-series data~\cite{liang2024bi,ahamed2024tscmamba,an2025damba,an2026dema}.
For example, TimeMachine~\cite{ahamed2024timemachine} harnesses Mamba to capture enduring patterns in multivariate time-series data, ensuring linear-complexity computation and minimal memory footprints for streamlined time-series processing.
Moreover, Mambaformer~\cite{xu2024integrating} combines selective SSM and Attention layers for the long- and short-term forecasting of weather, traffic flow, and more. 

\subsubsection{\textbf{Speech}}
Speech specifically refers to the vocalized form of human communication that involves vocalized expressions using specific phonetic sounds, words, grammar, and intonation patterns~\cite{abdel2014convolutional}. 
Recently, in the realm of speech-related tasks, researchers~\cite{chen2024rawbmamba} have made significant progress in developing Mamba-based models to tackle the emerging challenges encountered by existing model architectures, such as RNNs and Transformers. 
For example, SPMamba~\cite{li2024spmamba} and DPMamba~\cite{jiang2024dual} utilize bidirectional Mamba modules to capture a broader range of contextual information for speech separation, demonstrating a substantial improvement of 13\% in model performance and a 566\% reduction in computational complexity compared to a Transformer-based baseline when addressing speech separation tasks.
    
\subsubsection{\textbf{Motion}}
Human motion understanding and generation stand as a significant pursuit in a broad range of practical applications, including computer animation, game development, and robot manipulation.
However, semantic actions that occur infrequently within lengthy motion sequences make long-range motion modeling difficult. 
To address this issue, several studies have proposed the use of Mamba to capture spatiotemporal patterns in motion sequences~\cite{wang2024text}.
For instance, Motion Mamba~\cite{zhang2024motion} proposes a hybrid Mamba model, which leverages a hierarchical SSM layer to capture temporal patterns and introduces a bidirectional SSM layer to learn spatial knowledge, preserving motion consistency between frames.
Based on the comprehensive experiments, the Mamba-based model outperforms representative diffusion-based methods in human motion generation tasks, achieving a 50\% FID improvement and four times faster performance.
Additionally, MambaMOS~\cite{zeng2024mambamos} designs a motion-aware state space model that focuses explicitly on capturing variations in motion between consecutive time steps, which further emphasizes the exceptional capabilities of Mamba in achieving high-quality, lengthy sequence motion modeling.

\subsection{Non-Sequential Data}
Non-sequential data differs from sequential data in that it does not adhere to a specific order.
Its data points can be organized or accessed in any sequence without significantly impacting the data's meaning or interpretation~\cite{huang2011learning}.
This absence of inherent order presents difficulties for recurrent models such as RNNs and SSMs, specifically designed to capture temporal dependencies in data.
Surprisingly, Mamba, representing SSMs, has shown outstanding success in efficiently dealing with non-sequential data in recent developments.
In this section, we will review relevant studies about how Mamba effectively handles non-sequential data, including images, graphs, and point clouds. 

\subsubsection{\textbf{Image}}
As one of the most prevalent modalities, image data forms the foundation of various computer vision applications, e.g., face recognition, medical vision~\cite{yue2024medmamba}, and remote sensing~\cite{chen2024rsmamba,chen2024changemamba}.
Drawing inspiration from the success of Mamba in sequence modeling, there exists an intriguing opportunity to transfer this accomplishment from text processing to image analysis.
It involves treating an image as a series of patches, potentially paving the way for new avenues of exploration within the realm of computer vision. 
Thus, plenty of Mamba-based vision models have recently been developed to alleviate heavy computational resources and memory pressures while exhibiting competitive modeling capabilities~\cite{lin2024pixmamba,wang2024mamba,ruan2024vm,fu2024ssumamba}.
For example, Vision Mamba~\cite{zhu2024vision} incorporates bidirectional SSM to facilitate global visual semantic modeling and incorporates positional embeddings for location-aware visual comprehension. 
Not requiring attention mechanisms, Vision Mamba matches the modeling capacity of Vision Transformers while substantially decreasing computation time to subquadratic levels and upholding linear memory complexity.
Specifically, it outperforms the state-of-the-art baseline DeiT~\cite{touvron2021training} in terms of speed, being 2.8× faster, and also presents a remarkable reduction of 86.8\% in GPU memory usage during batch inference for feature extraction on high-resolution images (1248×1248).
Moreover, VMamba~\cite{liu2024vmamba} introduces 2D Selective Scan (SS2D) that serves as a bridge between 1D array scanning and 2D plane traversal, enabling Mamba to process visual data effectively.
Additionally, LKM-UNet~\cite{wang2024large} presents a Mamba-based model designed for medical image segmentation. This model leverages large Mamba kernels within a UNet architecture, showcasing superior performance in both local and global spatial modeling for visual inputs in medical imaging tasks.

\subsubsection{\textbf{Graph-structured Data}}
Graph modeling has found extensive utility in managing complex structures and relationships, including applications in domains like social networks~\cite{fan2019graph,fan2020graph}, recommender systems~\cite{fan2022graph}, and molecular interactions~\cite{huang2020skipgnn}.
Due to the powerful capabilities of Mamba in long-range modeling and high efficiency, several pioneering investigations have embraced the selective State Space Model (SSM) for non-sequential graph data~\cite{liu2024mamba4rec}.
These studies utilize state space models to encode context through hidden states during recurrent scans, allowing for input flow control, which resembles attention sparsification on graphs, presenting a data-dependent node selection process within graph modeling contexts~\cite{ye2021sparse}.
Moreover, Mamba is anticipated to enhance model efficiency during large-graph training tasks.
For example, Graph-Mamba~\cite{wang2024graph} introduces a novel Mamba-based block as a foundational component for graph modeling.
This block combines a graph flattening mechanism with the selection mechanism offered by Mamba, transforming sub-graphs into node sequences and facilitating input-dependent context filtering, respectively.
In a recent work, \citet{behrouz2024graph} propose a Graph Mamba Network (GMN), a new graph neural network format based on selective SSMs.
The authors reformulate the selective SSM into a graph learning format and provide theoretical justification for the power of the proposed network.
By addressing the emerging challenges in crucial steps of graph message passing, GMNs achieve remarkable performance in various aspects, surpassing GNNs and Transformer-based models in multiple benchmark datasets with diverse graph scales.
Furthermore, \citet{huang2024can} introduce the Graph State Space Convolution (GSSC) as a systematic extension of SSMs tailored for graph-structured data.
Specifically, GSSC incorporates distance-based graph convolution kernels into the SSM cell, aiming at enhancing expressive power and capturing long-range dependencies.
Through assessments conducted on ten benchmark datasets, the study~\cite{huang2024can} underscores the potential of GSSC as a potent and scalable model for graph machine learning.

\subsubsection{\textbf{Point Cloud}}
Point cloud is a crucial modality in computer vision, with a multitude of practical applications across domains like robotics, autonomous driving, and augmented reality~\cite{guo2020deep}.
Unlike image processing and graph learning, the analysis of point clouds presents unique challenges stemming from point clouds' inherent irregularity and sparsity, a 3D non-structured data.
To tackle these challenges, notable advancements have been made in deep learning-based approaches, with particular emphasis on Transformer-based models~\cite{yu2022point}.
However, the complexity of attention mechanisms is quadratic, bringing significant computational cost, which is not friendly to low-resource devices.
Noted by the recent advances of State Space Models (SSMs) in handling 1D sequences (e.g., language and speech) and 2D data (e.g., image and graph), there have been efforts to extend the application of Mamba to 3D point clouds~\cite{yi2024mvgamba}.
In general, these Mamba-based methods for point cloud analysis employ a two-step process~\cite{han2024mamba3d,zhou20243dmambaipf}.
First, the point cloud data is tokenized into discrete tokens using specific scanning methods.
Then, Mamba is utilized to capture the underlying patterns within these tokens.
For instance, PointMamba~\cite{liang2024pointmamba} proposes a hierarchical scanning strategy to encode local and global information of 3D point cloud and then utilizes plain Mamba as the backbone to extract features from serialized point tokens without incorporating additional complex techniques.
Point Cloud Mamba~\cite{zhang2024point} incorporates Mamba as the foundational model backbone to significantly reduce memory usage, demonstrating comparable (or superior) performance compared to Transformer-based counterparts. 

\subsection{Multimodal Data}
Integrating multiple modalities, such as language (sequential data) and images (non-sequential data), offers valuable and complementary information for artificial intelligence perception and scene understanding.
Recently, there has been significant research attention on Multimodal Large Language Models (MLLMs) that inherit the advanced capabilities of LLMs~\cite{wu2023multimodal}, including powerful language expression and logical reasoning.
While Transformers have been the dominant approach in this field, Mamba has emerged as a strong competitor by demonstrating impressive performance in aligning mixed-source data and achieving linear complexity scaling in sequence length, which makes Mamba a promising alternative to Transformers for multimodal learning~\cite{yang2024shmamba,liu2024robomamba}. 
For example, \citet{qiao2024vl} propose VL-Mamba to explore the utilization of Mamba's efficient architecture for solving vision-language tasks, using the pre-trained Mamba model for language understanding and incorporating a connector module to align visual patches with language tokens.
\citet{wang2024text} propose Text-controlled Motion Mamba~\cite{wang2024text}, which leverages Mamba to dynamically capture global temporal information based on text queries to enhance human motion understanding.
Besides, Fusion-Mamba~\cite{dong2024fusion} and Sigma~\cite{wan2024sigma} have tried to fuse complementary information from different thermal, depth, and RGB modalities.

\section{Applications}
\label{se:app}
In this section, we introduce several notable applications of Mamba-based models. To provide a comprehensive overview, we categorize these applications into: \emph{Natural Language Processing}, \emph{Computer Vision}, \emph{Speech Analysis}, \emph{Chemistry}, \emph{Recommender Systems}, and \emph{Robotics and Autonomous Systems}.

\subsection{Natural Language Processing}
In the natural language processing domain, recently, some Mamba-based models have emerged as alternatives to Transformer-based models for language modeling~\cite{waleffe2024empirical, zhao2024cobra, anthony2024blackmamba, bronnec2024locost, lieber2024jamba, he2024densemamba,xu2024rankmamba}, especially in applications involving extensive contexts such as \emph{Question Answering Systems} and \emph{Text Summarization}.

\subsubsection{\textbf{Question Answering Systems}.}
Question Answering (QA) involves AI models comprehending, reasoning, and responding using extensive knowledge bases, enabling coherent and contextually rich conversations, widely applied in chatbots and virtual assistants. Incorporating context from previous interactions is crucial for accurately addressing follow-up questions in multi-turn conversations. However, existing models face challenges in inference speed and computational efficiency, particularly in complex reasoning tasks. This leads to significant memory use and computational overhead, which limits scalability and real-time application efficiency. 
To address these limitations, recent studies have explored Mamba-based models to improve long-term dialogue management in QA Systems. 
For instance, Mamba-Chat~\cite{haven2023mambachat} is the first chat language model utilizing a state-space framework. The model maintains and updates its understanding of dialogues by employing state space representations, ensuring context awareness.
Jamba~\cite{lieber2024jamba} strategically alternates between Transformer and Mamba layers, incorporating MoE to enhance model capacity while optimizing parameter utilization. In common-sense reasoning and reading comprehension tasks, Jamba achieves performance comparable to larger Llama-2 models but with fewer parameters, demonstrating efficiency and effectiveness. 
Similarly, DenseMamba~\cite{he2024densemamba} introduces a novel method to enrich the propagation of hidden information across layers in SSMs by selectively incorporating hidden states from shallow layers into deeper layers. Compared to traditional Transformer-based models, this preserves crucial fine-grained information for superior performance in question-answering tasks. 
Overall, integrating Mamba-based models shows promising potential to advance QA systems by improving dialogue management and enhancing performance in complex reasoning tasks.
 
\subsubsection{\textbf{Text Summarization.}}
Text summarization aims to condense long texts by preserving essential information. Maintaining coherence and relevance is crucial in this task. Transformer-based models often struggle with long-sequence dependencies, potentially compromising coherence and relevance. In contrast, Mamba-based models leverage robust long-sequence processing capabilities, making them well-suited for processing coherent and context-rich text. 
Their robust architecture allows them to excel in summarization tasks by accurately capturing and condensing the essence of extensive documents. 
For instance, LOCOST~\cite{bronnec2024locost}, based on state space models, processes significantly longer sequences than sparse attention models. 
In long document abstractive summarization, LOCOST achieves performance comparable to the highest-performing sparse transformers of equivalent dimensions while reducing memory usage by up to 50\% during training and 87\% during inference. 
Additionally, SAMBA~\cite{ren2024samba} integrates Mamba with sliding window attention, enabling selective sequence compression into recurrent hidden states while retaining precise memory recall through attention mechanisms. SAMBA achieves a throughput 3.73 times higher than Transformers when handling input lengths of 128K, showcasing superior performance in tasks requiring long-context summarization.

\subsection{Computer Vision}
In addition to NLP applications, Mamba-based models have shown potential in the computer vision domain, representative applications like \emph{Disease Diagnosis} and \emph{Motion Recognition and Generation}.

\subsubsection{\textbf{Disease Diagnosis.}}
In clinical practice, medical images and videos provide critical insights into the morphology of organs or tissues. Efficient analysis of biomedical objects, such as lesions in large-scale 2D/3D images or videos, significantly enhances disease diagnosis and clinical treatment. 
However, CNN-based models like UNet face challenges in handling long-range dependencies because of their restricted receptive fields. This challenge is intensified by the typically larger sizes and higher resolution of medical images than natural images. Meanwhile, Transformer-based algorithms are computationally intensive, limiting their practicality in resource-constrained clinical settings. To overcome these limitations, numerous studies have adopted Mamba-based models in real medical environments~\cite{ma2024u, ruan2024vm, wang2024weak,liao2024lightm}. For instance, U-Mamba~\cite{ma2024u} and SegMamba~\cite{xing2024segmamba} both integrate a hybrid CNN-SSM block, merging the local feature extraction capabilities of convolutional layers with the long-range dependency modeling offered by SSMs. 
This hybrid approach outperforms the existing models in tasks such as 3D segmentation of abdominal organs in CT and MR images, segmentation of instruments in endoscopy images, and segmentation of cells in microscopy images. 
Similarly, CMViM~\cite{yang2024cmvim} addresses challenges in Alzheimer's disease (AD) diagnostic imaging by leveraging masked Vim autoencoders and contrastive learning across modalities, achieving the best performance in AD diagnostic imaging classification. 
Additionally, ProMamba~\cite{xie2024promamba} specializes in polyp segmentation. By incorporating Vision-Mamba architecture and prompt technology, this model achieves higher accuracy and better generalization than previous methods. For dynamic medical object segmentation in videos, Vivim~\cite{yang2024vivim} effectively compresses long-term spatiotemporal representations across different scales into sequences using the Temporal Mamba Block. This approach demonstrates enhanced performance and computational efficiency in disease diagnosis, such as ultrasound breast lesions segmentation and polyp segmentation in colonoscopy videos.

\subsubsection{\textbf{Motion Recognition and Generation.}}
Motion recognition and generation are critical in motion monitoring~\cite{golestani2020human}, computer animation~\cite{siarohin2021motion}, game development~\cite{nasri2020semg}, and film production~\cite{wang20233d}. However, transformer-based models encounter challenges related to computational and memory demands, limiting their applicability in resource-constrained environments. 
Additionally, Transformers and GCN-based models struggle with effectively capturing long motion sequences and complex spatial-temporal patterns in videos and 4D point clouds.
Recent studies have explored the use of Mamba to address these challenges, leveraging its robust performance and lower computational demands~\cite{li2024harmamba,chaudhuri2024simba,zhang2024motion}. For instance, HARMamba~\cite{li2024harmamba} utilizes a bidirectional SSM architecture to process data from wearable sensors, significantly reducing computational load and memory usage while maintaining high accuracy in real-time human motion recognition. Similarly, Simba~\cite{chaudhuri2024simba} integrates Mamba within a U-ShiftGCN framework, effectively handling longer sequences and complex spatial-temporal interactions, achieving the best results in skeletal action recognition from videos. 
Furthermore, Motion Mamba~\cite{zhang2024motion} and InfiniMotion~\cite{zhang2024infinimotion} are both for motion generation. 
Specifically, Motion Mamba utilizes hierarchical temporal Mamba blocks for processing temporal data and bidirectional spatial Mamba blocks for handling latent poses, ensuring motion consistency across frames and enhancing motion generation accuracy within temporal frames.
InfiniMotion introduces the Motion Memory Transformer with Bidirectional Mamba Memory, improving the transformer's memory capabilities to efficiently generate continuous, long-duration human motions (up to one hour and 80,000 frames) without overwhelming computational resources.

\subsection{Speech Analysis}
Speech signals inherently consist of thousands of samples. While this broad temporal context provides rich acoustic features, it also poses significant computational demands. To process speech signals effectively, several Mamba-based models have been successfully employed in diverse speech applications, notably in \emph{Speech Separation and Tagging} and \emph{Speech Enhancement}.

\subsubsection{\textbf{Speech Separation and Tagging.}}
Speech separation involves isolating individual speech signals from a multi-speaker environment. It is critical for enhancing the intelligibility and quality of audio communications. Meanwhile, audio tagging or classification involves mapping audio samples to their corresponding categories. Both tasks depend on capturing short-range and long-range audio sequential patterns.
Although transformer-based models have been the leading architecture of these applications, they face significant challenges in quadratic computational and memory costs due to their self-attention mechanisms. 
Recently, there has been a shift toward employing state space models for speech separation~\cite{jiang2024dual,li2024spmamba} and audio tagging~\cite{zhang2024mamca,bhati2024dass}. Specifically, DPMamba~\cite{jiang2024dual} utilizes selective state spaces to capture dynamic temporal dependencies in speech signals, encompassing both short-term and long-term forward and backward dependencies. 
SPMamba~\cite{li2024spmamba} integrates the TF-GridNet model, replacing its transformer components with bidirectional Mamba modules. DASS~\cite{bhati2024dass} integrates knowledge distillation with state-space models, allowing for tagging sound events in audio files lasting up to 2.5 hours. Meanwhile, MAMCA~\cite{zhang2024mamca} focuses on Automatic Modulation Classification (AMC) by introducing the selective state-space model as its backbone, effectively addressing both accuracy and efficiency challenges associated with long-sequence AMC. By adopting state-space models, these models demonstrate a qualitative improvement, capturing a broader range of contextual information and enhancing overall effectiveness, thereby proving the superior scalability of SSMs in handling long durations.

\subsubsection{\textbf{Speech Enhancement.}} 
Speech enhancement (SE) aims to extract clear speech components from distorted signals, producing enhanced signals with improved acoustic characteristics. As a front-end processor, SE is pivotal in numerous speech applications, including assistive hearing technologies~\cite{kumar2022deep}, speaker recognition~\cite{bai2021speaker}, and automatic speech recognition~\cite{malik2021automatic}. Mobile audio devices face challenges due to limited resources. Recent studies have explored the application of Mamba, leveraging its powerful performance and reduced computational demands in SE tasks~\cite{quan2024multichannel,chao2024investigation,zhang2024mamba,shams2024ssamba}. For instance, TRAMBA~\cite{sui2024tramba} leverages a hybrid architecture combining Transformers and Mamba to improve speech quality for mobile and wearable platforms, specifically targeting acoustic and bone conduction. It achieves a remarkable tenfold reduction in memory consumption compared to the current leading models. Additionally, oSpatialNet-Mamba~\cite{quan2024multichannel} leverages Mamba for long-term multichannel speech enhancement, achieving outstanding results for static and moving speakers.

\subsection{\textbf{Chemistry}}
Protein design, molecular design, and genomic analysis are pivotal in advancing drug discovery and biotechnology~\cite{scott2016small,li2024empowering,zhou2025hd}.
Leveraging the Mamba-based model significantly reduces the complexities of modeling long sequences in these domains~\cite{brazil2024mamba,guo2024saturn, schiff2024caduceus}. Specifically, PTM-Mamba~\cite{peng2024ptm} and ProtMamba~\cite{sgarbossa2024protmamba} are protein language models based on the Mamba architecture. PTM-Mamba utilizes bidirectional gated Mamba blocks and structured state space models, efficiently processing long sequences while reducing computational demands.
ProtMamba is designed to be homology-aware yet alignment-free, adept at handling extensive contexts across hundreds of protein sequences. 
Both models maintain high efficiency and accuracy even with large data sets, providing critical tools for protein design. 
Meanwhile, generative molecular design aims to simulate molecules with tailored property profiles from a specific distribution. However, current models lack the efficiency required to optimize high-fidelity oracles, directly resulting in low success rates.
Saturn~\cite{guo2024saturn}, applying the Mamba architecture, utilizes its linear complexity and computational efficiency to surpass 22 competing models in drug discovery. Furthermore, understanding genomes is crucial for gaining insights into cellular biology. Challenges in genomic modeling include capturing interactions between distant tokens, considering the impacts of both upstream and downstream regions, and ensuring the complementarity of DNA sequences. Caduceus~\cite{schiff2024caduceus} and MSAMamba~\cite{thoutam2024msamamba}, both leveraging the Mamba model, excel in addressing these challenges. Caduceus, a DNA foundation model, enhances the Mamba architecture with BiMamba and MambaDNA components for bi-directional modeling and ensuring reverse complement equivariance, significantly outperforming existing models in long-range genomic tasks. 
Similarly, MSAMamba addresses the limitations of transformer-based models for DNA multiple sequence alignments by implementing a selective scan operation along the sequence dimension. This design extends the training context length of previous methods by eightfold, allowing a more comprehensive analysis of extensive DNA sequences.
Another illustration is SMILES-Mamba~\cite{xu2024smiles}, a system that pre-trains and fine-tunes Mamba models to forecast the absorption, distribution, metabolism, excretion, and toxicity characteristics of small-molecule drugs.

\subsection{Recommender Systems}
Recommender Systems are widely utilized in e-commerce~\cite{zhang2024linear,zhou2018micro,chen2023fairly} and social networks~\cite{fan2019deep_dscf,fan2019deep_daso,fan2018deep}, aiming to capture users' evolving preferences and the interdependencies among their past behaviors~\cite{zhao2024recommender,fan2022graph,qu2025diffusion}.
Although transformer-based models have demonstrated effectiveness in recommender systems~\cite{sun2019bert4rec}, they face computational efficiency challenges because of the quadratic complexity of attention mechanisms, especially when dealing with longer sequences of behaviors. Recently, several Mamba-based models have been applied to analyze long-term user behavior for personalized recommendations~\cite{yang2024uncovering,liu2024mamba4rec,wang2024echomamba4rec,su2024mlsa4rec,cao2024mamba4kt}. 
For example, Mamba4Rec~\cite{liu2024mamba4rec} pioneers the use of selective state space models for efficient sequential recommendation, enhancing model performance while maintaining inference efficiency. Similarly, RecMamba~\cite{yang2024uncovering} explores Mamba's effectiveness in lifelong sequential recommendation scenarios (i.e., sequence length $\ge$ 2k), achieving comparable performance to benchmark models while cutting down training time by 70\% and reducing memory costs by 80\%. Furthermore, EchoMamba4Rec~\cite{wang2024echomamba4rec} integrates a bidirectional Mamba module with frequency-domain filtering to accurately capture intricate patterns and interdependencies within user interaction data. It demonstrates superior performance over existing models, delivering more precise and personalized recommendations. Additionally, Mamba4KT~\cite{cao2024mamba4kt} is designed explicitly for knowledge tracing in intelligent education, leveraging the Mamba model to capture enduring correlations between exercises and student knowledge levels.
As educational datasets expand, this method suggests a promising avenue for enhancing prediction accuracy, model efficiency, and resource utilization in knowledge tracing research.
{More recently, SSD4Rec~\cite{qu2024ssd4rec} proposes a novel pure-SSM structure for efficient sequential recommendation, which harnesses the advanced nature of Structured State-Space Duality to enable attention-like parallel computation on item sequences while scaling linearly with the input lengths.
SSD4Rec allows variable-length input without any padding/truncation and achieves comprehensive long-range user behavior modeling from bi-directional perspectives.}

\subsection{Robotics and Autonomous Systems} 
The main goal of robotics and autonomous systems is to develop models capable of comprehending visual environments and performing intricate actions. Multimodal Large Language Models (MLLMs) currently used in robotics face significant challenges in two primary aspects: 1) limited capacity for handling intricate tasks requiring advanced reasoning, and 2) substantial computational expenses with fine-tuning and inference tasks. 
Due to their advantages in inference speed, memory utilization, and overall efficiency, Mamba models are emerging as a promising foundation for autonomous and intelligent systems~\cite{cao2024mamba,liu2024robomamba,jia2024mail}, promising superior performance and substantial scalability potential. For example, RoboMamba~\cite{liu2024robomamba} integrates a vision encoder with Mamba to create an end-to-end robotic MLLM. This method aligns visual data with language embeddings by co-training, enhancing the model with visual common sense and robot-specific reasoning while ensuring efficient fine-tuning and inference capabilities. Similarly, \citet{jia2024mail} introduce MaIL, an imitation learning (IL) policy architecture that uses Mamba as a backbone. MaIL bridges the gap between efficiency and performance in handling sequences of observations. Extensive evaluations of real robot experiments demonstrate that MaIL provides a competitive alternative to traditional, large, and complex Transformer-based IL policies.

\section{Challenges and Opportunities}
\label{se:future_direction}
The preceding sections have thoroughly surveyed the latest advanced techniques and varied applications associated with Mamba.
However, the studies of Mamba are still in their nascent stages, and there exist considerable challenges and opportunities ahead.

\subsection{Mamba-based Foundation Models}
By scaling up the model sizes to the billion level over large-scale mixture-of-source corpora, foundation models (FMs) exhibit impressive zero-shot learning capabilities, which have enabled FMs to excel in a wide range of general tasks~\cite{bommasani2021opportunities,dai2025how,ning2025survey}.
As a representative example, recent years have witnessed the booming success of Transformer-based large language models, especially ChatGPT, motivating a growing enthusiasm for exploring foundation models in various domains.
Even though Transformers are the main drivers of the success, they suffer from pressing computation and memory efficiency issues~\cite{tay2022efficient}, which come with the exponentially growing training memory proportional to the attention-based model size and the laborious auto-regressive decoding during inference.
In response to these issues, a promising alternative backbone, i.e., Mamba~\cite{gu2023mamba,mamba2}, for foundation models has recently emerged. 
% a new class of sequence models based on state space models.
Mamba offers the content-aware learning capabilities of Transformers while scaling the computation linearly with input length, making it effective in capturing long-range dependencies and enhancing efficiency in both training and inference.
Given these advantages, developing Mamba-based foundation models for specific domains holds great potential, which offers an opportunity to address the issues faced by Transformer-based models.

\subsection{Hardware-Awareness Computation}
Foundation models, characterized by their large sizes and intensive matrix operations like matrix multiplications and convolutions, require cutting-edge hardware such as GPUs and TPUs for high-throughput training and inference.
These advanced hardware enable researchers to work with larger datasets and achieve state-of-the-art performance across various domains. 
Nonetheless, the existing foundation models still fall short of fully exploiting the computational capabilities of the hardware, resulting in limited model efficiency~\cite{tay2022efficient}.
As a promising alternative for enhancing computation efficiency, Mamba-1~\cite{gu2023mamba} and Mamba-2~\cite{mamba2} put forth hardware-aware computation algorithms, namely the Parallel Associative Scan and the Block-decomposition Matrix Multiplication.
These algorithms take into account the inherent characteristics of GPUs and TPUs, including factors such as message transmission between devices, offering a fresh perspective on addressing the computation efficiency problem.
Inspired by this, exploring novel hardware-efficient algorithms, such as FlashButterfly~\cite{fu2023simple}, to optimize hardware utilization offers a promising avenue for conserving resources and accelerating computation, benefiting not only SSMs but also other architectures like Transformers and RNNs.

\subsection{Trustworthy Mamba Models}
The development of SSMs is expected to bring significant benefits to various industries, including e-commerce, healthcare, and education.
At the same time, being a data-dependent model like many existing architectures, Mamba models could pose severe threats to users and society~\cite{marques2022delivering}.
These threats arise from several factors, such as erratic decision-making, privacy concerns, and more.
Therefore, ensuring trustworthiness in Mamba models is essential across four critical dimensions~\cite{liu2022trustworthy}: \emph{Safety\&Robustness}, \emph{Fairness}, \emph{Explainability}, and \emph{Privacy}.

\subsubsection{\textbf{Safety\&Robustness}}
Large foundation models have been proven to be highly vulnerable to adversarial perturbations, which can jeopardize the safety and robustness of these models when deployed in safety-critical applications~\cite{wei2024jailbroken,ning2024interpretation,fan2023adversarial}.
Meanwhile, Mamba-based models are not exempt from these vulnerabilities~\cite{malik2024towards}.
In the pursuit of being a reliable alternative to Transformer, it is essential to investigate and enhance the safety and robustness of Mamba-based models. 
To be specific, the model outputs should be robust to small perturbations in their inputs. 
One potential solution could involve automatically pre-processing prompts before feeding them into Mamba-based models. Additionally, as a representative technique, adversarial machine training~\cite{huang2011adversarial} can be employed to enhance the safety and robustness of Mamba models.

\subsubsection{\textbf{Fairness}}
Large foundation models, trained on extensive datasets, tend to be unintentionally exposed to the biases and stereotypes present in the extensive training corpus~\cite{ma2024fairness}, which can manifest in the generated outputs.
For instance, within the domain of LLMs, the biases can lead to discriminatory responses influenced by user profile attributes like gender and age, reinforcing stereotypes and unfairly treating specific user groups~\cite{jiang2024item}.
While recent efforts have been made to address the issue of fairness in LLMs, there is still a gap in research regarding the non-discrimination and fairness of Mamba models.
Thus, further exploration and study are necessary to bridge this gap.

\subsubsection{\textbf{Explainability}}
Deep learning models have often been criticized for their "black-box" nature, and the explainability of deep learning models has emerged as a popular topic in the research community, which indicates the capacity to comprehend and interpret the decisions or predictions generated by a model~\cite{dovsilovic2018explainable}.
By explaining model predictions, users can make more informed decisions based on the model's outputs. 
To this end, several techniques have been proposed to provide plausible innate explanations for neural architectures based on attention mechanism~\cite{hu2023seat}.
Moreover, researchers have investigated the capabilities of Transformer-based language models to generate natural language descriptions to explain their answers~\cite{yuan2024back}.
Although an increasing number of studies have attempted to take full advantage of Mamba, studies on comprehending the functioning of Mamba models are still at an early stage, and further investigation is still needed.

\subsubsection{\textbf{Privacy}}
The protection of privacy builds trust between users and Mamba-based models.
When users have confidence that their privacy is respected, they are more likely to engage with the AI systems, share relevant information, and seek assistance without fear of misusing their data.
Thus, this trust is vital for the widespread adoption and acceptance of Mamba models. 
One effective strategy for mitigating privacy risks involves cross-verifying the output of Mamba models and screening sensitive content~\cite{kim2024propile}. 
Moreover, Federated Learning is poised to bolster privacy during the training of Mamba models, wherein the model is trained on numerous decentralized edge devices or servers housing local data samples, without data exchange. This methodology aids in preserving the localization and privacy of the data. 
Furthermore, integrating privacy-conscious regularization techniques such as differential privacy constraints during training shows promise in preventing overfitting on sensitive data.

\subsection{Applying Emerging Techniques from Transformer to Mamba}
The Transformer, being the dominant backbone, has led the AI community to develop numerous unique tools aimed at enhancing the performance of attention-based models.
Fortunately, by connecting SSMs and attention, the SSD framework introduced by Mamba-2~\cite{mamba2} allows us to develop a shared vocabulary and library of techniques for Transformer and Mamba.
In light of this, an important future direction arises, i.e., to explore how the emerging techniques designed for Transformer-based models can be effectively applied to Mamba-based models.

\subsubsection{\textbf{Parameter-efficient Finetuning}}
Large foundation models, scaling up their parameters to billions, have witnessed groundbreaking advancements in multiple fields. 
Nevertheless, their extensive scale and computational requirements present significant challenges when tailoring them for specific downstream tasks.
To this end, several parameter-efficient finetuning (PEFT) techniques, including the LoRA~\cite{hu2021lora} and Adapter families~\cite{gao2024clip,karimi2021compacter}, have been proposed, which involve minimizing the adjustment of parameters or the need for extensive computational resources during finetuning. 
Drawing inspiration from the recent achievements in employing PEFT for large language models constructed using Transformer layers, the adoption of PEFT for Mamba models has emerged as an intriguing topic, with the goal of broadening their range of applications in downstream tasks. 
For instance, the deployment of LoRA (Low-Rank Adaptation) is anticipated to facilitate rapid finetuning for the SSD models, thus enabling the widespread application of Mamba across various domains.
However, the specifics of implementing these PEFT techniques for Mamba-based models are yet to be determined and require further investigation.

\subsubsection{\textbf{Catastrophic Forgetting Mitigation}}
Catastrophic forgetting, also known as catastrophic interference, refers to the phenomenon observed in machine learning models where they experience a significant loss in performance on previously learned tasks when trained on new tasks~\cite{kemker2018measuring}.
This issue poses a challenge for foundation models because they need to retain knowledge from pre-training tasks and demonstrate consistent performance across different downstream domains.
As a promising architecture of the foundation model, Mamba necessitates a thorough investigation to address catastrophic forgetting issues.
Recent research has suggested resolving this challenge by encapsulating task-specific needs through Reward Maximization and Distribution Matching strategies~\cite{korbak2022reinforcement,korbak2022controlling}.
Moreover, continual learning methods have also been developed to mitigate catastrophic forgetting in Transformer-based language models~\cite{wang2022continual,kar2022preventing}.
These techniques can also be applied to Mamba models by connecting SSMs and attention, but remain under-explored.

\subsubsection{\textbf{Retrieval-augmented Generation (RAG)}}
Being among the most sophisticated techniques in AI, RAG can provide dependable and current external knowledge, offering significant utility for a multitude of tasks~\cite{lewis2020retrieval,ding2024survey}.
Large Language Models have recently showcased groundbreaking language comprehension and generation capabilities, despite encountering inherent limitations like hallucinations and outdated internal knowledge. 
In light of RAG's potent capacity to offer current and valuable supplementary information, Retrieval-Augmented LLMs have emerged to leverage extraneous knowledge databases for enhancing the generative quality of LLMs~\cite{chen2024benchmarking}. 
RAG can be incorporated with Mamba models to assist them in producing high-quality outputs, which is a promising future research direction.

{
\subsection{Benchmarking Mamba Variants across Diverse Domains}
With the rapid development of Mamba-related research, there is an emerging need for thorough and comprehensive benchmarking of Mamba and its variants across diverse tasks and datasets~\cite{park2024can,liu2024vision}, particularly the comparisons with the Transformer-based counterparts.
Such benchmarking will provide important insights into the application of Mamba models.
Currently, a few pioneering studies have contributed in this direction.
For example, Mamba-in-Vision~\cite{rahman2024mamba} provides an extensive comparison of Mamba, CNN, and Transformer models in five typical visual tasks, namely image classification, object detection, semantic segmentation, action classification, and remote sensing.
This investigation demonstrates the promising capabilities of Mamba models, showing that they can achieve competitive performance while using fewer parameters and FLOPs.
Furthermore, \citet{waleffe2024empirical} investigate whether Mamba models can perform on par with Transformers when evaluated on large-scale benchmarks spanning a variety of natural language tasks.
Their findings indicate that, although pure SSM-based models often equal or outperform Transformers on several tasks, both Mamba and Mamba-2 models tend to underperform compared to Transformers in scenarios that demand strong copying abilities, in-context learning, or long-context reasoning.
Additionally, \citet{han2025demystify} provide a meticulous analysis of the advantages and disadvantages of Mamba models compared to linear attention across six aspects: input gate, forget gate, shortcut, absence of attention normalization, single-head design, and modified block design.
Their empirical evaluation on vision tasks highlights the forget gate and block design as the primary contributors to Mamba’s success, while the other four design elements are found to be less critical.
Despite these efforts, significant gaps remain in benchmarking Mamba models, not only within the visual and language domains but also in other important areas such as graph learning and time-series modeling. Future research should place greater emphasis on domain-specific evaluation metrics and real-world performance considerations to fully realize the potential of Mamba models.
}

\subsection{Inherent Limitations remain in Mamba}
Despite Mamba's popularity for achieving attention-like performance with linear-time computational scaling, it inherits fundamental limitations from its underlying structured state space model (SSM) architecture.
Most existing Mamba models are inherently causal as underscored by their recurrent view of computation, i.e., Equations~\eqref{eq:ssm1} and \eqref{eq:ssm2}, which confines their strong performance primarily to autoregressive settings and restricts their applicability in non-autoregressive tasks~\cite{hwang2024hydra}.
For example, one study demonstrates that generalized state space models, including LSTMs and Mamba, fail at the classic \emph{Copying} task: they cannot accurately reproduce sequences longer than the size of their latent state~\cite{jelassi2024repeat}.
This failure stems from a core memory trade-off: while transformers employ input-dependent memory that scales with sequence length (albeit less memory-efficiently), SSMs utilize a compact, fixed-size state.
This design likely sacrifices the ability to reliably retrieve and repeat arbitrary segments of the input context.
On the other hand, another critical analysis indicates that discrete SSMs like S4 and Mamba struggle with \emph{State-Tracking} problems, such as permutation composition~\cite{merrill2024illusion}.
These capabilities are essential for core LLM functions like tracking narrative entities, playing chess from notation, or evaluating code.
This difficulty may stem from the fact that Mamba relies on unbounded depth to approximate recurrent neural networks (RNNs), leading it to encounter state-tracking limitations analogous to those of transformers.
Addressing these mathematical and empirical challenges represents a significant opportunity for impact across the broader machine learning community.
\section{Conclusion}
\label{se:conclusion}
Mamba, an emerging deep learning architecture, has demonstrated remarkable success across diverse domains, such as language generation, image classification, recommendation, and drug discovery, owing to its powerful modeling capabilities and computational efficiency. 
Recently, increasing efforts have been made to develop Mamba-based deep learning models with more powerful representation learning capabilities and lower computation complexity. 
Given the rapid advancement of Mamba, there arises an urgent demand for a systematic overview.
To bridge this gap, in this paper, we provide a comprehensive review of Mamba, focusing on its architecture advancements, data adaptability, and application areas, offering researchers both an in-depth understanding and an overview of the latest developments in Mamba.
Additionally, given that Mamba research is still in its nascent stages, we also discuss current limitations and present promising directions for future investigation.

\begin{acks}
The research described in this paper has been partially supported by the General Research Funds from the Hong Kong Research Grants Council (project no. PolyU 15207322, 15200023, 15206024, 15224524, and 15216225), Hong Kong Research Grants Council’s Theme-based Research Scheme (No.T43-513/23-N), Hong Kong Research Grants Council’s Research Impact Fund (No.R1015-23), Hong Kong Research Grants Council’s Collaborative Research Fund (No.C1043-24GF), Internal research funds from Hong Kong Polytechnic University (project no. P0042693, P0048625, P0051361, P0052406, P0059586, and P0052986), and Sheertek International (HK) Limited. This work was supported by computational resources provided by The Centre for Large AI Models (CLAIM) of The Hong Kong Polytechnic University.
\end{acks}

\bibliographystyle{ACM-Reference-Format}
\bibliography{references}

\end{document}